\documentclass{article}

\usepackage[preprint]{neurips_2019}

\usepackage[utf8]{inputenc} 
\usepackage[T1]{fontenc}    
\usepackage{hyperref}       
\usepackage{url}            
\usepackage{booktabs}       
\usepackage{amsfonts}       
\usepackage{nicefrac}       
\usepackage{microtype}      
\usepackage{easy-todo}

\usepackage{graphicx, xcolor, graphbox, amsmath, array}
\usepackage{arydshln, lscape}

\newcommand{\eq}[1]{\begin{equation} #1 \end{equation}}

\title{Latent feature disentanglement for 3D meshes}

\author{%
  Jake Levinson \\
  Google Research \\
  \texttt{jlev@google.com} \\
  \And
  Avneesh Sud \\
  Google Research \\
  \texttt{avneesh@google.com} \\
  \And
  Ameesh Makadia \\
  Google Research \\
  \texttt{makadia@google.com} \\
}

\begin{document}

\maketitle

\begin{abstract}
  Generative modeling of 3D shapes has become an important problem due to its relevance to many applications across Computer Vision, Graphics, and VR. In this paper we build upon recently introduced 3D mesh-convolutional Variational AutoEncoders which have shown great promise for learning rich representations of deformable 3D shapes. We introduce a supervised generative 3D mesh model that disentangles the latent shape representation into independent generative factors. Our extensive experimental analysis shows that learning an explicitly disentangled representation can both improve random shape generation as well as successfully address downstream tasks such as pose and shape transfer, shape-invariant temporal synchronization, and pose-invariant shape matching.
\end{abstract}

\section{Introduction}
The ability to generate new 3D shapes is a fundamentally important objective for many applications, especially in Virtual Reality where the availability of large collections of varied 3D assets is necessary to create rich virtual environments. Much of the recent progress in this area has been facilitated by the introduction of new large scale shape datasets such as ShapeNet (\cite{shapenet}) and Dynamic FAUST (\cite{dfaust:CVPR:2017}), which have made viable the approaches based on data-driven deep learning techniques (see for example \cite{3dgan,achlioptas18icml,bib:meshvae,tan18cvpr}). For many important applications it is not random shape generation that is desired but rather some user-controlled generation: the ability to manipulate an object by its parts (e.g. \cite{borosan12rig,dubrovina19arxiv}), or transfer pose characteristics across deformable shape instances (e.g. \cite{sumner04sig,Gao:2018:AUS:3272127.3275028:DeformationTransfer}). These goals require a generative model that disentangles the underlying factors of variation in the data.

Learning disentangled representations is a well-studied problem in machine learning (\cite{schmidhuber92neural,bengio13pami,locatello18disentangled,achille18jmlr,factorvae,hinton11icann,infogan}). In the most general context, the explanatory or generative factors are \emph{a priori} unknown, so the goal of disentanglement is to learn latent factors that are mutually independent and that capture maximal variation in the data. Unsupervised approaches, however, make it difficult to control the interpretation of the disentangled factors. Indeed, many natural modes of variation such as shape and color may be highly correlated in training data, even when they describe semantically independent features.

In this work, we introduce a new generative model for 3D shapes that explicitly disentangles the shape representation by its observable generative factors. Our model builds upon the generative Variational AutoEncoders (\cite{vae}) which have shown promising results for learning rich representations of deformable 3D meshes categories, e.g. humans, animals (\cite{bib:meshvae,tan18cvpr}). The model is trained on combinations of synthetic and real datasets where the variations of interest can be controlled during mesh generation. This allows us to generate large scale datasets with the necessary supervision (our model knows when training shapes share a latent factor). In addition to a dataset of articulated cylinders, we show results on a large scale dataset of approximately 3M human shapes exhibiting extreme pose and shape variation (following \cite{varol17_surreal}).\footnote{Although a parametric model is used to generate our training data, our model is agnostic to this (it only sees the 3D meshes) and can thus scale trivially to non-parametric shape datasets.} Our evaluations include an analysis of the model's latent disentanglement properties and experiments for several downstream applications: shape and pose transfer, temporal synchronization, and pose-independent shape matching.  

Along with disentanglement, we improve the core performance of the basic Mesh VAE by incorporating a distortion-sensitive loss term that promotes more realistic shape generation, and an alternative technique for latent sampling that can overcome overparameterization of latent spaces (since the optimal latent dimensionality is typically unknown). One insight from our experiments is that a disentangled model can outperform a vanilla model with the same base architecture and generative capacity. This validates the hypothesis that disentangled models learn compact, robust representations.

One surprise in our results is that certain training modes lead to models that are disentangled from the generative standpoint, but not for inference -- i.e., the latent representation itself is `entangled', but the generator learns to disregard redundant or irrelevant latent information. Our primary model, however, is disentangled for both use cases.
\section{Related Work}
In contrast to the unsupervised disentangling models discussed above, for learning from visual data latent factors are often observable and in some way explicitly supervised. Model training may exploit temporal structure (e.g. in videos, \cite{denton17nips,villegas17iclr}), or generation of synthetic data with controlled latent factors (\cite{bib:inverse_graphics_network,worrall2017iccv,yang2015weaklysupervised}). Our approach most closely relates to the Inverse Graphics Network (\cite{bib:inverse_graphics_network}) which manipulates factors of variation within training mini-batches. This approach requires knowing which generative factors are being varied, but does not require supervision of the explicit parametric transformations as in \cite{worrall2017iccv} and \cite{yang2015weaklysupervised}.

A number of recent works explore learning disentangled generative models (Variational AutoEncoders, \cite{vae}, Generative Adversarial Networks, \cite{gan}) where the latent representation is decomposed into an observed (potentially interpretable) component and a component for the remaining variability (\cite{kingma_semisup_vae,disentangled_vae,de2018dgpose,mathieu16nips}). In these approaches, the interpretable latent factors (e.g. class label or human pose) typically require direct supervision with a regression or classification loss.

While there are many generative models for 3D data such as volumes, point clouds, and meshes (\cite{3dgan,achlioptas18icml,tan18cvpr,bib:meshvae}), disentangled models, in particular generative models, are an under-explored area. Recently \citet{dubrovina19arxiv} learns a part-aware factorized embedding space. Shapes can be generated by manipulating object parts, but the model generates volumetric shapes. 

In addition to the works described above, it is important to note disentangled representations have been explored for numerous applications related to image data. Although a full review is out of scope here, to highlight different applications we refer the reader to topics on face images (\cite{liu2018exploring,tran17cvpr,NeuralFace2017,shu18eccv}), intrinsic image decomposition (\cite{barrow78cvs,barron15pami}), and characteristic transfer across images (e.g. motion, \cite{DBLP:journals/corr/abs-1808-07371-dancenow}, appearance, \cite{Zanfir_2018_CVPR}, and domain, \cite{DBLP:journals/corr/abs-1901-10024:PuppetGAN}).

\paragraph{Articulated shape models.}
There is a significant body of work in representation learning for deformable articulated 3D shapes, notably of humans and animals. There are several parametric human shape models that capture the intrinsic human shape variation (\cite{Anguelov:2005:SSC:1186822.1073207, Allen:2006:LCM:1218064.1218084, yang2014semantic, pishchulin17pr}). Such approaches align a human mesh template to a set of 3D human scans, such as CAESAR (\cite{caesar1999}), and compute the principal components on mesh vertex displacements or transformation matrices. To represent various human pose shapes, parametric skeleton skinning based approaches and deformation based approaches have been used. Skinning based approaches such as SMPL (\cite{SMPL:2015}) and \cite{Allen:2006:LCM:1218064.1218084} compute vertex positions from the body pose using learnt skinning weights. Deformation-based approaches such as  SCAPE (\cite{Anguelov:2005:SSC:1186822.1073207}),~\cite{freifeld2012lie},~\cite{hasler2009statistical, hirshberg2012coregistration} use various representations of deformations to a reference mesh. More recently, rotation invariant (\cite{gao2016efficient}) and as-consistent-as-possible (\cite{gao2017sparse}) deformation features have been used in mesh convolutional neural networks to extract a deformation embedding (\cite{tan18cvpr}) and perform unpaired shape deformation transfer using 3D shape CycleGAN (\cite{Gao:2018:AUS:3272127.3275028:DeformationTransfer}). In contrast, our work focuses on explicit shape and pose latent feature disentanglement for general articulated meshes. To capture a natural distribution of human poses, several 3D human animation datasets have been collected. SURREAL (\cite{varol17_surreal}) performs SMPL fits to CAESAR shapes and activities from ~\cite{cmu_mocap} using ~\cite{Loper:SIGASIA:2014}. \cite{Dyna:SIGGRAPH:2015} and \cite{dfaust:CVPR:2017} provide direct scans from humans performing various activities. 
Finally, there is work on capturing 3D shapes of animals, including parametric deformable models (\cite{cashman2012shape, Zuffi:CVPR:2018}), and part based representations (\cite{ntouskos2015component}).
\section{Generating 3D Shapes}

\subsection{Variational autoencoding}
Variational autoencoders (VAEs) are a widely-used framework for generative modeling. A VAE assumes that data $x$ is jointly distributed with certain latent variables $z$, which are typically given an independent Gaussian prior, $z \sim \mathcal{N}(0, I)$. To infer $z$ from $x$, we model the posterior distribution $p(z|x)$ by an encoder $\mathrm{Enc}(z|x; \theta)$, which we take to be a neural network. Similarly, we model the likelihood $p(x|z)$ by a decoder network $\mathrm{Dec}(x|z; \theta)$, which allows the model to be used generatively. Training a VAE consists of approximately minimizing the KL divergence of the estimated posterior $\mathrm{Dec}(x|z; \theta)$ from the true posterior $p(x|z)$, by maximizing the so-called Evidence Lower-Bound (ELBO). For more on VAE training, see \cite{vae}.

\subsection{MeshVAE and the disentangled model}

We based our model on the mesh variational autoencoder (MeshVAE) of \cite{bib:meshvae} (in principle our contributions could be incorporated into any similar Mesh VAE  model e.g. \cite{tan18cvpr}). The MeshVAE acts on input data consisting of per-vertex features on a mesh, i.e. an input is $\mathbf{m} \in \mathbb{R}^{|V| \times f}$, where there are $|V|$ vertices and $f$ features (for us, $f=3$, the vertex coordinates). The model outputs global latent parameters $z \in \mathbb{R}^n$. The architecture relies crucially on the mesh topology and is entirely convolutional, except for a single initial (fully-connected) decoding layer mapping the latent encoding to a set of per-vertex hidden features.

The architecture is as follows:
\begin{enumerate}
    \item[(i)] The {\bf encoder} uses feature-steered mesh convolutions (FeaStNet, see \cite{bib:feastnet}), followed by mean-pooling along vertices. We model the posterior $p(z|\mathbf{m})$ as a Gaussian, so that the encoder gives a latent mean and variance $\mathrm{Enc}(\mathbf{m}) = (\mathbf{\mu}(\mathbf{m}), \mathbf{\sigma}(\mathbf{m}))$ in $\mathbb{R}^{n_s + n_p}$, where $n_s$,  $n_p$ correspond to the number of shape and pose features.

    \item[(ii)] For VAE training, we sample a {\bf latent feature} $\mathbf{x} = (\mathbf{s}, \mathbf{p}) \sim \mathcal{N}(\mu, \sigma)$, consisting of a {\bf shape feature} $\mathbf{s} \in \mathbb{R}^{n_s}$ and a {\bf pose feature} $\mathbf{p} \in \mathbb{R}^{n_p}$. At inference, we simply use $\mathbf{x} = \mu$.
    
    \item[(iii)] The {\bf decoder} generates per-vertex hidden features from one fully-connected layer, then applies a sequence of FeaStNet convolutional layers.
\end{enumerate}

\subsection{MeshVAE-D: Training for disentanglement}

A baseline MeshVAE produces an `entangled' latent encoding, which affords little or no control in shape generation. The goal for the disentangled model, MeshVAE-D, is for the latent features $(\mathbf{s}, \mathbf{p})$ to capture shape and pose separately, and we took three steps to this end.

{\bf Batching}. We structured the training set (SMPL) into doubly-supervised training batches, allowing us to train the model while fully supervising the desired factors of variation. We first structured the dataset into pairs of meshes, with each pair having either the same underlying body shape (i.e. subject identity) or the same pose (cf. the supervision techniques in \cite{bib:inverse_graphics_network,worrall2017iccv,yang2015weaklysupervised}). Each training batch then consisted of $N$ shape-constant or pose-constant pairs of meshes. For Faust shapes (\cite{dfaust:CVPR:2017}), pose labels are not available, so we only used shape-constant batches. Notably, despite only having access to partial supervision on Faust, the trained model successfully extracts pose and shape features from Faust meshes and is able to conduct pose transfer (see Fig. \ref{fig:interpolate plus transfer}).

{\bf Clamping}. For a pair of meshes from a training batch, the encoder produces latent features $(\mathbf{s}_1, \mathbf{p}_1)$ and $(\mathbf{s}_2, \mathbf{p}_2)$. During training, for shape-constant pairs, we replaced the latent shape vectors by their joint mean $\bar{\mathbf{s}} := \tfrac{1}{2}(\mathbf{s}_1 + \mathbf{s}_2)$ before passing them to the decoder. For pose-constant pairs, we instead clamped the latent pose vectors to $\bar{\mathbf{p}} := \frac{1}{2}(\mathbf{p}_1 + \mathbf{p}_2)$.

{\bf Latent variance loss}. We added a loss term $L_{\text{latent}}$ equal to the within-batch variance in the clamped latent feature: $\tfrac{1}{2}||\mu_{\mathbf{s},1} - \mu_{\mathbf{s},2}||^2$ for shape-constant pairs and $\tfrac{1}{2}||\mu_{\mathbf{p},1} - \mu_{\mathbf{p},2}||^2$ for pose-constant pairs.

Our clamping approach is similar to \cite{bib:inverse_graphics_network}, which not only averaged the latent features but stops gradients from passing through the clamped neurons. With the latter approach, since the pose encodes much more information than the body shape, it becomes necessary to train with a higher proportion (5-to-1) of shape-constant (i.e., pose-varying) batches. We found that stopping gradients had a mild negative impact on model performance, so our model does not do it.

\subsection{Loss and regularizers}
The VAE training loss consisted of two terms: $L^2$ reconstruction error $L_{\text{recon}}$, plus the KL divergence loss term $L_{\text{KL}}$ of the latent mean and variance from a Gaussian $\mathcal{N}(\mathbf{0}, I)$. For disentanglement, we included the latent variance loss $L_{\text{latent}}$ defined above. As an additional regularization to improve surface smoothness, we introduced a geometric distortion loss based on \cite[Eq. (3)]{bib:distortion}:
\eq{ \label{eq:distortion}
L_{\text{distortion}} = \frac{1}{|E|}\sum_{\substack{\text{edges} \\ i - j}} ||\Delta \mathbf{v}_i - \Delta \mathbf{v}_j||^2,
}
where $\Delta \mathbf{v} = \mathbf{v}^{\text{out}} - \mathbf{v}^{\text{in}}$ is the reconstruction displacement at $\mathbf{v}$. The distortion loss penalizes distortion between neighboring vertices, apart from a common translation relative to the base mesh. The resulting meshes are more realistic in terms of both surface texture and fine detail (see Fig. \ref{fig:distortion and shape recog}); moreover, generated meshes from this model retain smoothness even as the generated shape variation goes beyond the range of shapes seen during training (see Fig. \ref{fig:random_decodings}). In sum, the loss function was
\begin{equation*}
L = L_{\text{recon}} + \alpha L_{\text{KL}} + \beta L_{\text{latent}} + \gamma  L_{\text{distortion}}
\end{equation*}
and we used $\alpha = 10^{-8}, \beta = 10^0, \gamma = 3 \cdot 10^1$. 
\subsection{Models}
We compared our model to the following baselines: (1) an unmodified Mesh VAE, (2) model trained directly to do pose transfer between meshes, and (3) a model based on latent feature permutation during training (MeshVAE-P).

\subsection{Transfer model baseline}
We trained the Transfer model directly on a pose transfer task. We constructed a dataset of triples $(\mathbf{m}_{\text{target}}, \mathbf{m}_{\text{diff\_subject}}, \mathbf{m}_{\text{diff\_pose}})$ taken from SMPL, where the second and third meshes have, respectively, the same pose (but a different subject) and the same subject (but a different pose) as the target. The model is shown $(\mathbf{m}_{\text{diff\_subject}}, \mathbf{m}_{\text{diff\_pose}})$ and asked to predict $\mathbf{m}_{\text{target}}$. We used an architecture similar to the MeshVAE, with two encoders, one for shape and one for pose, with respectively $\tfrac{1}{2}$ and $\tfrac{3}{4}$ the hidden layer widths compared to the full model. The dimensions of the latent space and decoder were left unchanged, and we did not clamp the latent vectors.

\subsection{Permute model baseline (MeshVAE-P)}
We trained the Permute model without clamping and variance loss, instead permuting latent features of batch pairs during training. That is, we swap $\mathbf{s}_1$ and $\mathbf{s}_2$ in a shape-constant batch (or $\mathbf{p}_1$ and $\mathbf{p}_2$ in a pose-constant batch) before passing to the decoder. By construction, the exchanged latent features still describe the same true meshes, so the decoder learns to reconstruct the same output mesh. MeshVAE-P produces decoded meshes of similar quality to MeshVAE-D, but the latent features themselves are poorly disentangled: the shape vector ends up carrying pose information -- in fact \emph{more} pose information than shape information (see \ref{fig:histograms}) -- which the decoder learns to ignore (see \ref{fig:mve}). This baseline highlights a key distinction between \emph{generative} disentanglement (possible using MeshVAE-P or MeshVAE-D) and \emph{inferential} disentanglement (only possible with MeshVAE-D). Indeed, MeshVAE-P model performs closer to the baseline on an inference task related to shape.
\section{Experiments}

\subsection{Articulated cylinders}

We first trained our model on a toy dataset consisting of meshes shaped as cylinders with a single bend of angle $\theta \in [40^\circ, 180^\circ]$ (1 pose parameter) and varying arm lengths $\ell_1 \in [0.5, 2.5], \ell_2 \in [0.5, 2.5]$ and radius $r \in [0.1, 0.2]$ (3 shape parameters), see Fig. \ref{fig:cylinder}. For train/test splits, we held out a range of values for each parameter (see appendix).

\begin{figure}
    \centering
    \includegraphics[align=c, height=0.9in]{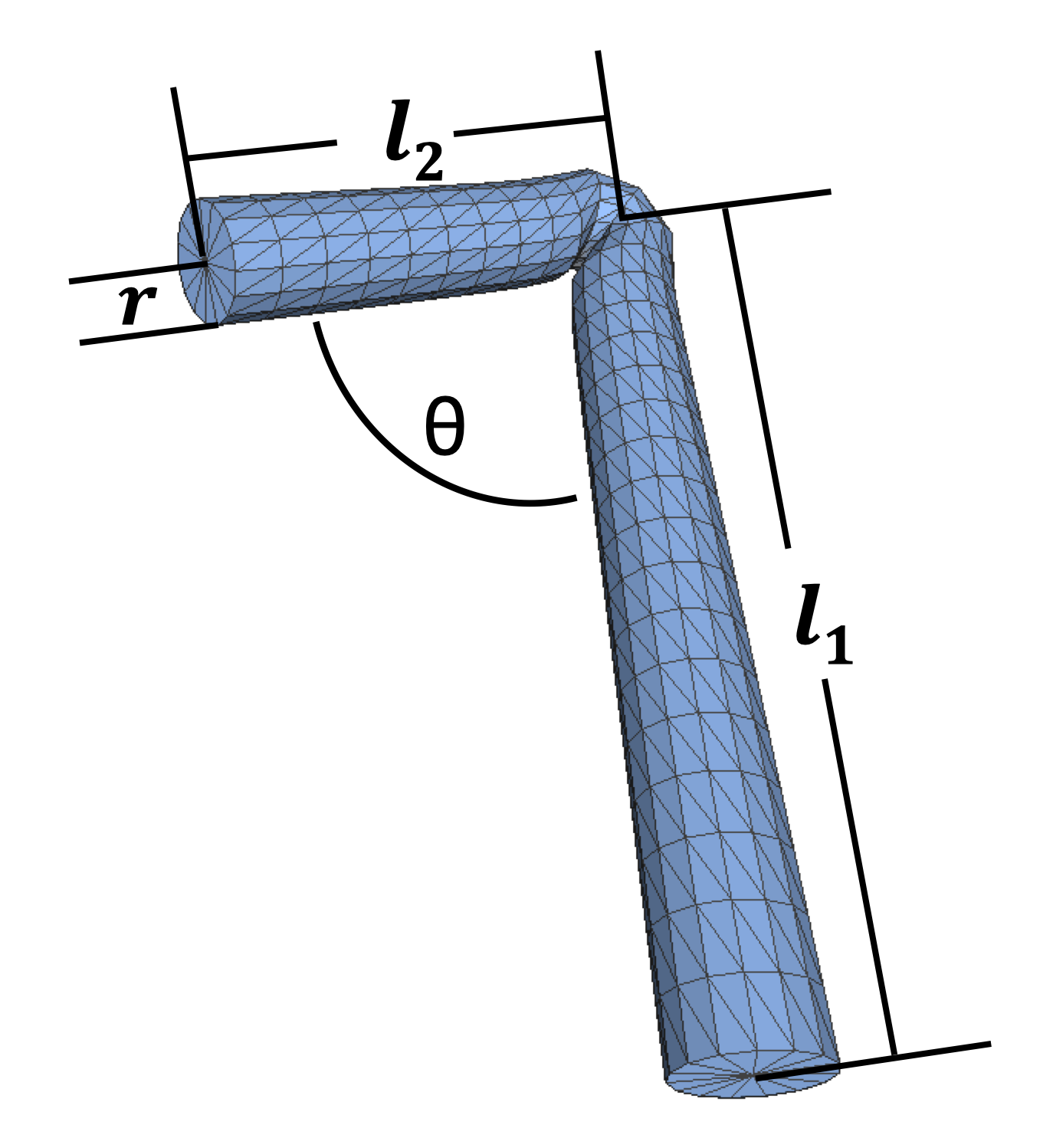}
    \qquad
    \begin{tabular}{|c|cccc|} \hline
        Parameter & $\theta_{\text{GT}}$ & $l_1{}_{\text{GT}}$ & $l_2{}_{\text{GT}}$ & $r_{\text{GT}}$ \\ \hline
        $\theta_{\text{est}}$ & 0.9628 & -0.0146 & -0.0065 & 0.0215 \\
        $l_1{}_{\text{est}}$ & -0.0192 & 0.9992 & - & - \\
        $l_2{}_{\text{est}}$ & -0.0047 & - & 0.9953 & - \\
        $r_{\text{est}}$ & -0.0034 & - & - & 0.9801 \\ \hline
    \end{tabular}
    \caption{\small {\bf Left}: An articulated cylinder. {\bf Right}: Pearson's correlation between pose ($\theta$) and shape $(l_1, l_2, r)$ parameters for ground truth (columns) and MeshVAE-D reconstructions (rows, using estimated parameters). Cross-correlations for pairs of shape features are omitted.}
    \label{fig:cylinder}
\end{figure}

The resulting models successfully disentangle the cylinder shape from the pose angle. We performed pose and shape exchanges by swapping latent features between cylinders with different shape and pose parameters, then recovered the latent parameters $\theta, \ell_1, \ell_2, r$ using a least squares fit of cylinders to corresponding vertex positions of the decoded shapes. We computed Pearson's correlation between the ground truth and estimated parameters, finding successful semantic disentanglement, with strong correlations between latent features and weak correlation across latent features.

\subsection{Human shapes} \label{subsubsec:human shapes}
Next, we trained a model on a combined dataset of human shapes. We combined shapes from the parametric mesh dataset SMPL and shapes from Faust, which consists of motion-captured meshes, labeled by subject identity but not pose.
Training batches consisted of 8 pairs of meshes, alternating between SMPL and Faust in a ratio of 5:5:1 (SMPL shape-constant batches : SMPL pose-constant batches : Faust batches). Note that Faust batches are always shape-constant because Faust does not have pose labels.

For train/test splits, we held out two subjects and one activity from Faust; for SMPL, we held out 100 pose sequences and all subjects whose leading four shape parameters fell within distance $1$ of the points $(\pm1, \pm1, \pm1, \pm1) \in \mathbb{R}^4$ ; the SMPL shape distribution overall was sampled uniformly from $-3$ to $3$ in each parameter. This is a much broader distribution than used by SURREAL (\cite{varol17_surreal}) which samples from the unit Gaussian. This was necessary to generate a dataset with extreme variations in human shape. We will provide all the sampled shape parameters and details so the test and train datasets can be reproduced exactly.

For direct reconstruction of input meshes, the model shows improved vertex error relative to a baseline Mesh VAE (Fig. \ref{fig:mve}) and compared to a model trained directly on pose transfer.

\begin{figure}
    \centering
    \begin{tabular}{|c|c|} \hline
        Model & MVE (cm) \\ \hline
        MeshVAE & 3.6 \\ \hline
        MeshVAE-D {\bf (Ours)} & {\bf 2.7} \\ \hline
        MeshVAE-P (Permute) & 2.8 \\ \hline
        Transfer model & 3.7 \\ \hline
    \end{tabular}
    \begin{tabular}{|c|c|} \hline
    Model & MVE (cm) \\ \hline
    MeshVAE & n/a \\ \hline
    MeshVAE-D {\bf(Ours)} & $4.0 \pm 3.6$ \\ \hline
    MeshVAE-P (Permute) & $3.8 \pm 3.7$ \\ \hline
    Transfer model & $7.4 \pm 9.3$ \\ \hline
    \end{tabular}

    \caption{\small {\bf Left}: Mean vertex error for direct mesh reconstruction on SMPL test set. {\bf Right}: Mean vertex error across 500 examples in the pose transfer experiment.}
    \label{fig:mve}
\end{figure}

\subsection{Encoder disentanglement}
We first assessed the quality of the latent encoding itself. We stored the latent shape and pose encodings for all meshes, then calculated the $L^2$ distance between latent pose encodings for: (a) pairs of meshes in the same pose, (b) pairs of meshes of the same shape, and (c) random pairs of meshes.

\begin{figure}
    \includegraphics[width=1\linewidth]{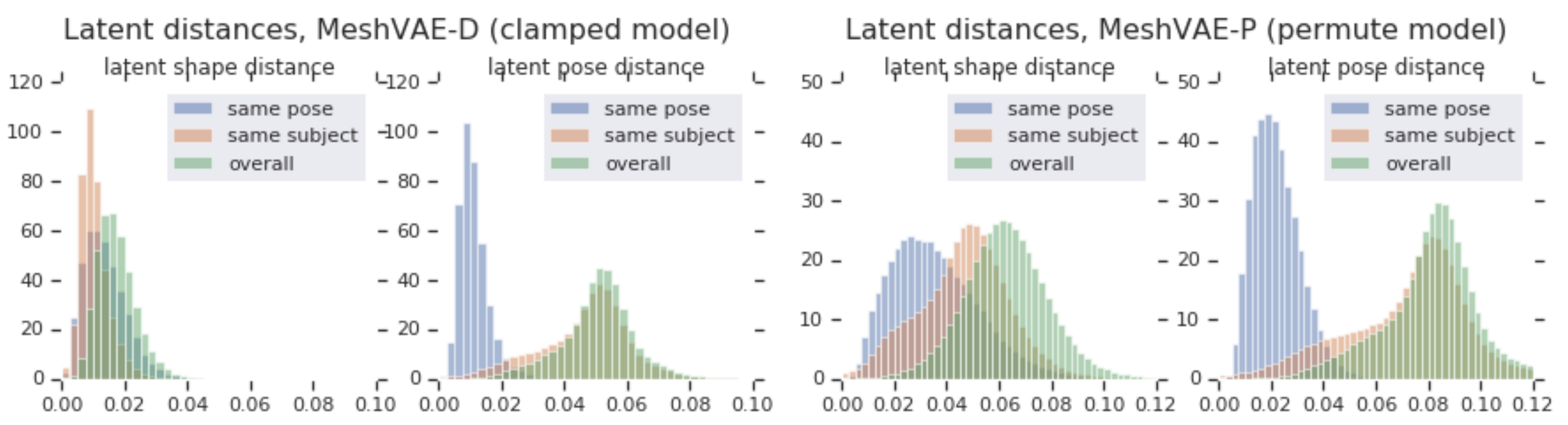}
    \caption{\small Distribution of latent distance between encodings of pairs of meshes with either the same shape, the same pose, or neither (random pairs). Distance in $\mathbb{R}^d$ is scaled by $\tfrac{1}{\sqrt{d}}$.}
    \label{fig:histograms}
\end{figure}

In a perfectly disentangled model, the distances (a) should be zero, while (b) and (c) should have similar distributions of latent distances. In practice, we instead observe $0 < (\text{a}) \ll (\text{b}) \approx (\text{c})$ for MeshVAE-D. We then repeated the calculation with latent shape encodings, where we expect the reverse, i.e. $0 < (\text{b}) < (\text{a}) \approx (\text{c})$. The distribution of distances is shown in the histograms in Fig. \ref{fig:histograms}, showing good disentanglement for our clamped model MeshVAE-D. By contrast, MeshVAE-P, trained by permuting rather than clamping the latent features, is poorly disentangled: the latent shape encoding is more responsive to pose than to shape! In particular, latent shape proximity in this model is more indicative of pose alignment ($(\text{a}) < (\text{b})$) than shape similarity.

\subsection{Decoder disentanglement}
To assess disentanglement on the level of the decoder, we attempted to generate shapes while holding one or the other feature fixed.

For fixed-pose, variable-shape generation, we took a pose encoding $\mathbf{p}$ from a fixed real mesh and generated shape encodings $\mathbf{s} \sim \mathcal{N}(0, \sigma),$ where $\sigma$ was the observed latent scale across the dataset. For random pose generation, since our model overparametrizes the true pose, the latent pose distribution does not, in practice, occupy the entire latent (pose) space -- making it a challenge to generate suitable random poses. To get around this problem, we examined the model's latent pose distribution using a modified PCA, computed at inference from the pose encodings of 120k random training meshes. The top 80 principal components account for $>99.9\%$ of the latent pose distribution variance. We then generated Gaussian random pose vectors $\mathbf{p}$ within these principal axes (weighted by the singular values) and combined them with a fixed shape encoding $\mathbf{s}$ from a fixed mesh from the test set. See Fig. \ref{fig:random_decodings}. A more detailed analysis of performance relative to latent dimensions is in the appendix.

Note that the level of variation (particularly in body shape) in the generated meshes goes beyond that of the training set. We view these results as compound benefits of having both a \emph{disentangled} and \emph{geometrically-based} model: we are able to vary the mesh in a controlled way, while our geometric priors, such as the distortion term \eqref{eq:distortion}, ensure that variation in vertex predictions is smoothed out locally to form plausible mesh deformations, rather than just degrading the mesh (see Fig. \ref{fig:distortion and shape recog} right).

\begin{figure}[h]
    \centering
    \begin{tabular}{cc}
    \includegraphics[height=2.2cm]{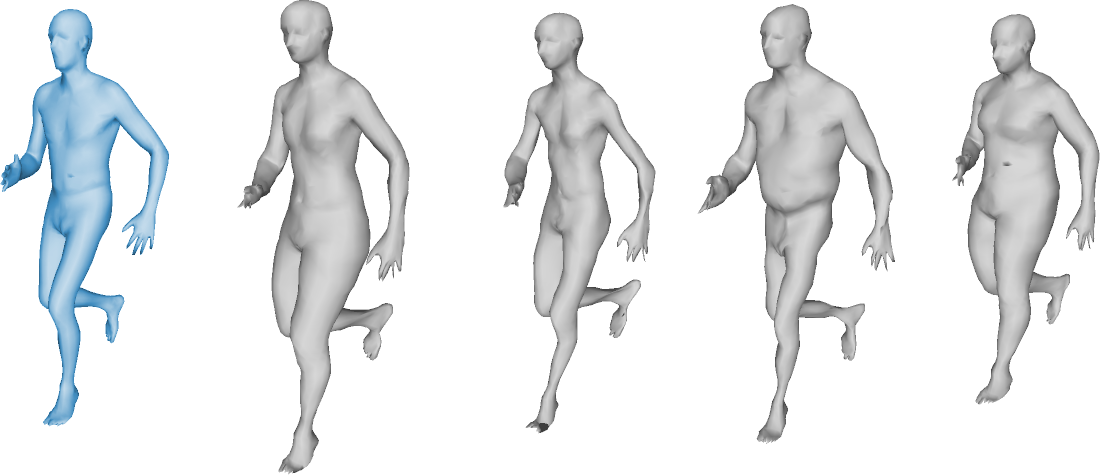} &
    \includegraphics[height=2.2cm]{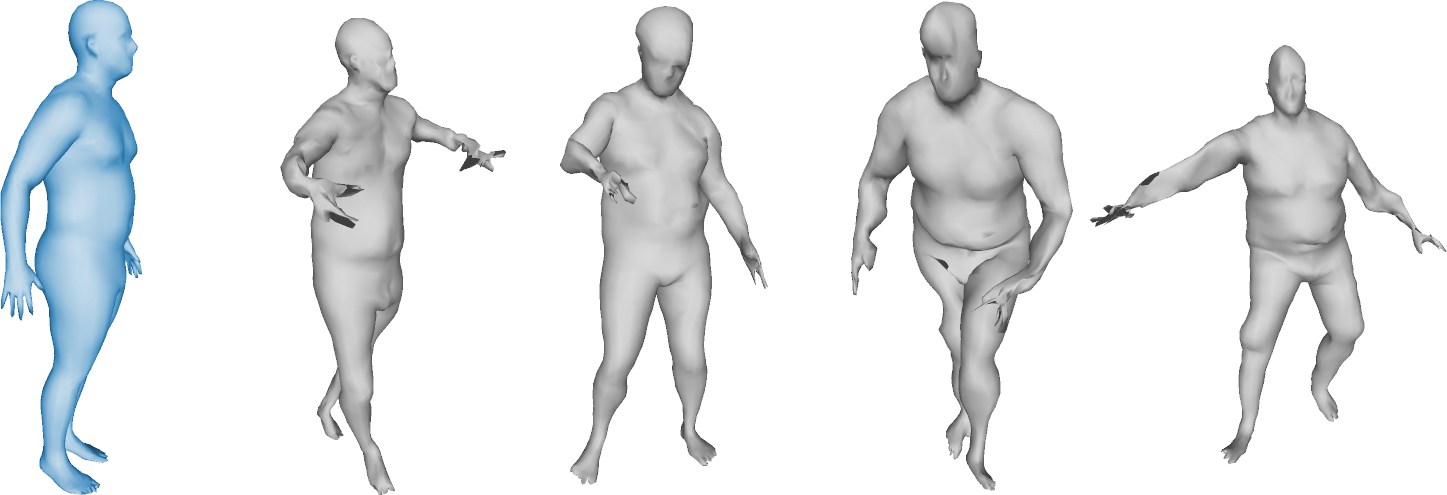} \\
    \includegraphics[height=2.2cm]{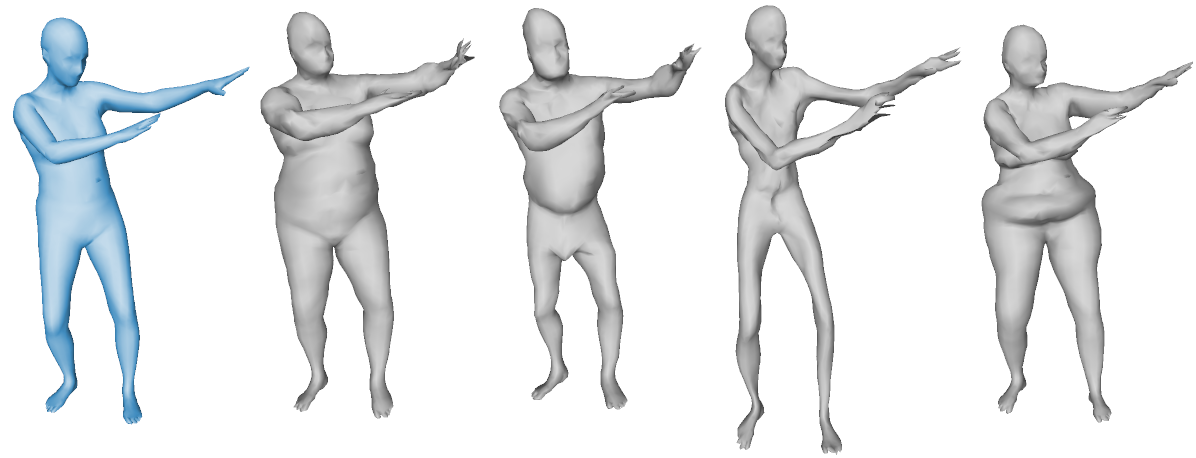} &
    \includegraphics[height=2.2cm]{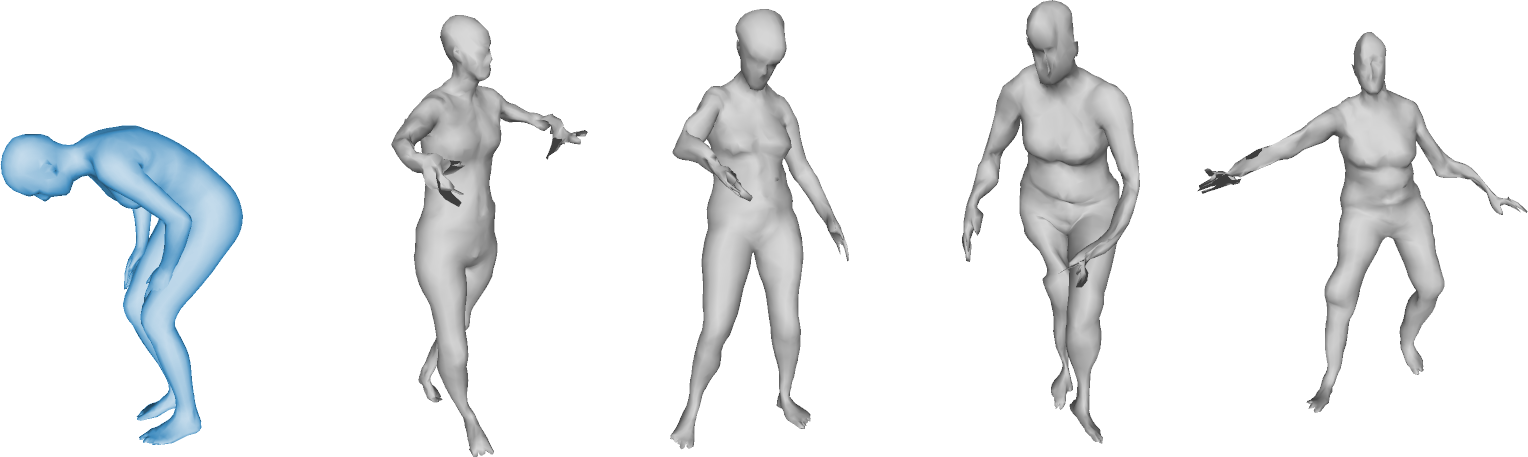}
    \end{tabular}
    \caption{\small Meshes from Gaussian random shape vectors and fixed real pose encoding (left column) or vice versa (right column). Our method can generate plausible shapes well outside the SMPL distribution.}
    \label{fig:random_decodings}
\end{figure}
\begin{figure}
    \begin{tabular}{cc}
    \begin{tabular}{c|c|c|c}
    Model & top1 & top2 & top3 \\ \hline
    MeshVAE & 31.8 & 39.4 & 43.8 \\
    MeshVAE-D {\bf (Ours)} & \bf 47.8 & \bf 57.6 & \bf 63.2 \\
    MeshVAE-P (Permute) & 46.0 &  55.3 & 59.9
    \end{tabular} &
    \begin{tabular}{ccc}
    \includegraphics[width=0.2\linewidth]{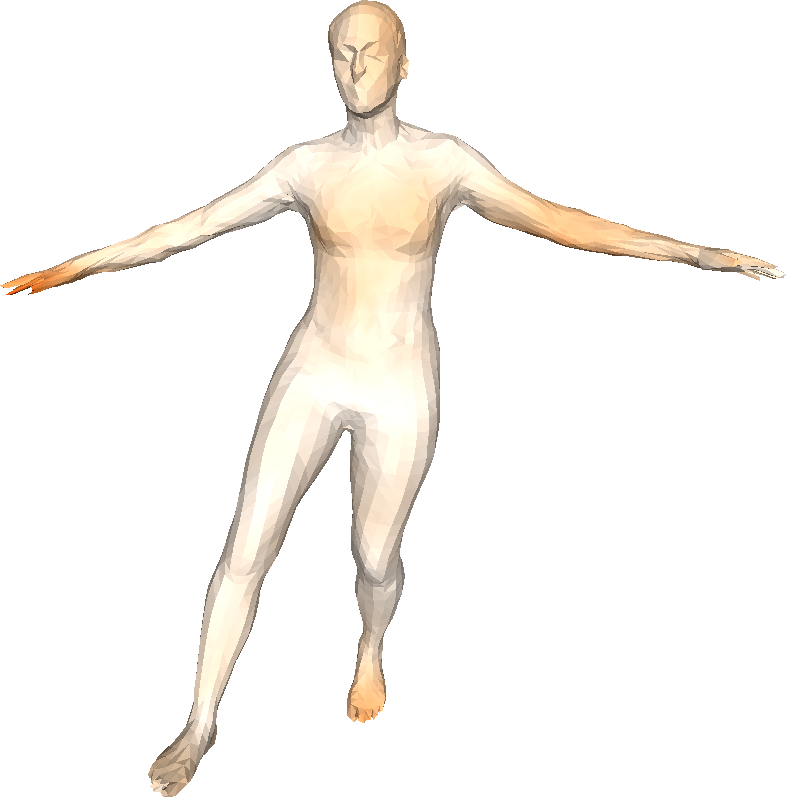} &
    \includegraphics[width=0.2\linewidth]{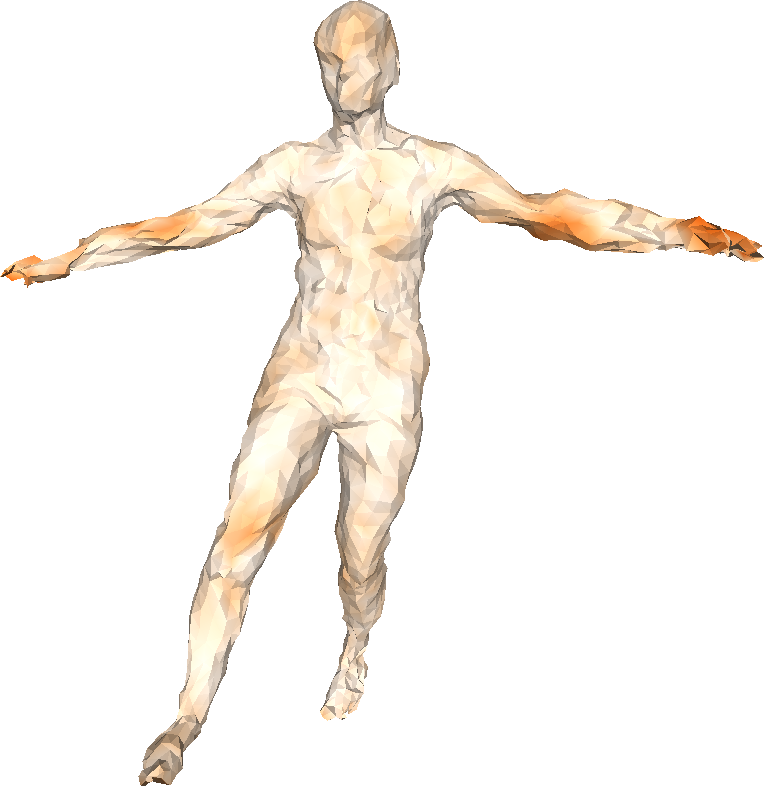}
    \end{tabular}
    \end{tabular}
    \caption{\small {\bf Left}: Top-$k$ score on shape recognition task. {\bf Right}: Impact of the distortion loss term on reconstructed mesh quality. Left is MeshVAE-D, right standard MeshVAE.}
    \label{fig:distortion and shape recog}
\end{figure}

\subsection{Pose and shape transfer}
A primary application of a disentangled model is to transfer poses and body shapes from one mesh to another. We used the dataset of triples $(\mathbf{m}_{\text{target}}, \mathbf{m}_{\text{diff\_subject}}, \mathbf{m}_{\text{diff\_pose}})$ constructed for the pose transfer baseline (see Sec. \ref{subsubsec:human shapes}). While the baseline model is trained directly on this task, the disentangled model instead produces $\mathbf{m}_{\text{target}}$ by combining the appropriate latent attributes from $\mathbf{m}_{\text{diff\_subject}}$ and  $\mathbf{m}_{\text{diff\_pose}}$. See Fig. \ref{fig:interpolate plus transfer}.

To ensure that the task presented a challenge for the model, we tested on a subset of triples, requiring the secondary meshes to have extrinsic mean vertex distance from $\mathbf{m}_{\text{target}}$ of at least $5$cm for $\mathbf{m}_{\text{diff\_subject}}$ and $15$cm for $\mathbf{m}_{\text{diff\_pose}}$. Surprisingly, the primary model outperformed the model trained directly on triples (Fig. \ref{fig:mve}).

\subsection{Pose synchronization on Faust}
\begin{figure}[h]
    \centering
    \includegraphics[width=1.0\linewidth]{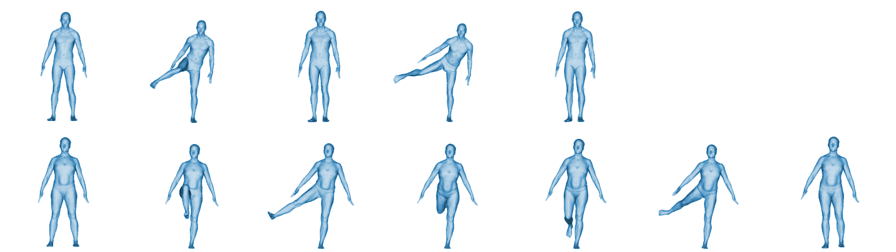} \\
    \includegraphics[width=1.0\linewidth]{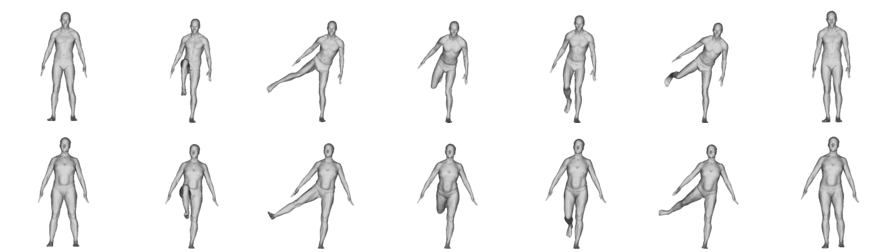} \\
    \caption{\small Pose synchronization based on the latent pose encoding. {\bf Top:} Selected frames from original motion sequences (580 and 1250 frames). {\bf Bottom:} Selected frames from dynamically synchronized sequences (1251 frames).}
    \label{fig:synchronize}
\end{figure}

Next, we evaluated the model on a pose synchronization task using dynamic time warping (DTW, \cite{dtw}). Given two sequences $\mathbf{m}^{(i)}, \mathbf{n}^{(j)}$ and costs (energies) $d(\mathbf{m}^{(i)}, \mathbf{n}^{(j)}),$ DTW produces a sequence of pairs $(i_1, j_1), (i_2, j_2), \ldots$ such that $i_1 \leq i_2 \leq \ldots$, $j_1 \leq j_2 \leq \cdots$, and every $\mathbf{m}^{(i)}$ is matched to at least one $\mathbf{n}^{(j)}$ and vice versa, minimizing the total energy. We performed the synchronization task using cost given by $L^2$ distance between latent pose encodings for $\mathbf{m}^{(i)}$ and $\mathbf{n}^{(j)}$.
This task is especially interesting on Faust because the sequences consist of similar motions (jumping jacks, running on the spot, and so on) but are not synchronized and do not have pose labels. See Fig. \ref{fig:synchronize}.

\subsection{Shape recognition on SMPL}
As a complementary experiment to pose synchronization, we performed a shape recognition task: we selected $N=100$ subjects at random from the test set, and for each subject chose two random meshes $\mathbf{m}_i, \mathbf{n}_i$. We then had the model predict, for each $\mathbf{m}_i$, which of the $N$ counterparts $\{\mathbf{n}_i\}_{i=1}^N$ comes from the same subject. We used nearest-neighbors assignment based on the shape encoding only (MeshVAE-D, MeshVAE-P) or the overall latent encoding (MeshVAE-D). Our model significantly outperforms the baseline on this task (Fig. \ref{fig:distortion and shape recog}), correctly identifying nearly half the meshes (top-1 score) and 2/3 (top-3 score). We also observe a moderate improvement over MeshVAE-P, for which the latent representation is only partially disentangled.

\subsection{Latent pose interpolation}
\begin{figure}[h]
    \begin{tabular}{ccc|c|c||ccccc}
    $\Delta_{\text{pose}}$ & $\Delta_{\text{subj}}$ & target & swap & direct &
    $t=0$ & $t=0.25$ & $t=0.5$ & $t=0.75$ & $t=1$ \\ \hline
    
    \includegraphics[align=c,scale=0.06]{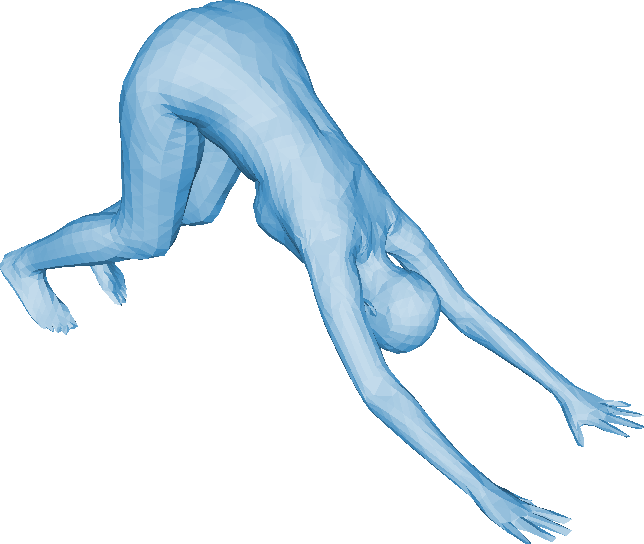} &
    \includegraphics[align=c,scale=0.06]{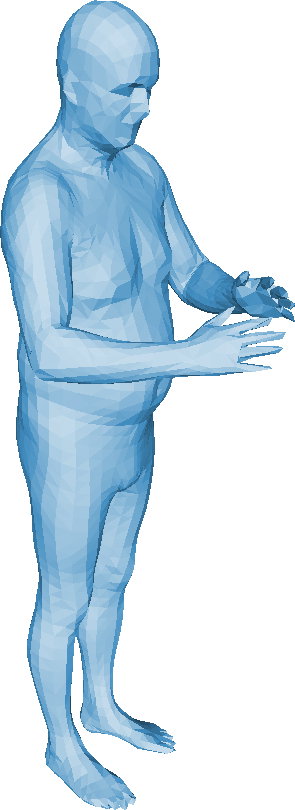} &
    \includegraphics[align=c,scale=0.06]{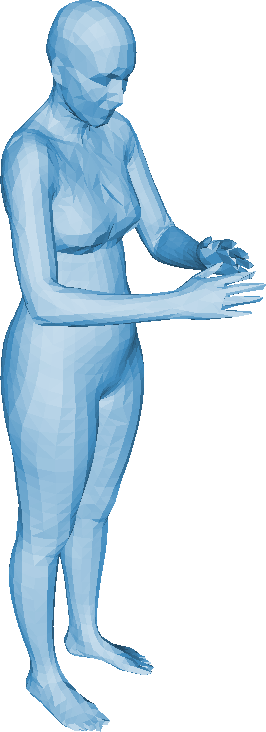} &
    \includegraphics[align=c,scale=0.06]{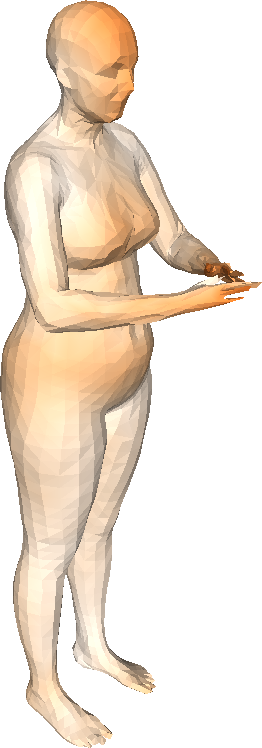} &
    \includegraphics[align=c,scale=0.06]{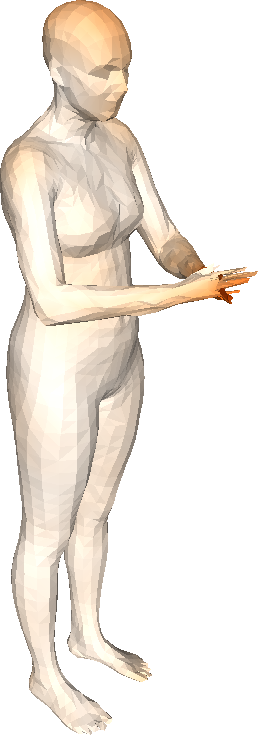} &
    \includegraphics[align=c,scale=0.10]{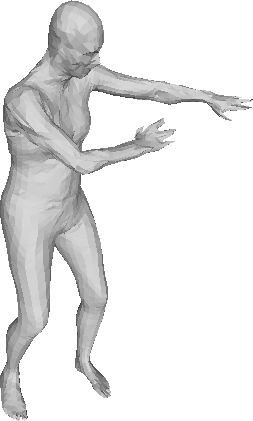} &
    \includegraphics[align=c,scale=0.10]{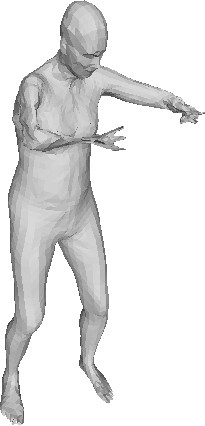} &
    \includegraphics[align=c,scale=0.10]{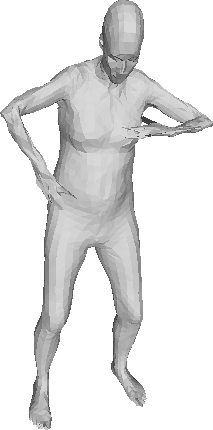} &
    \includegraphics[align=c,scale=0.10]{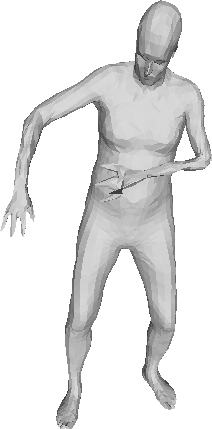} &
    \includegraphics[align=c,scale=0.10]{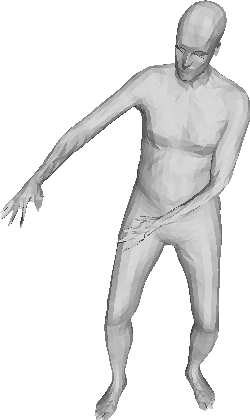} \\ \hline
    
    \includegraphics[align=c,scale=0.06]{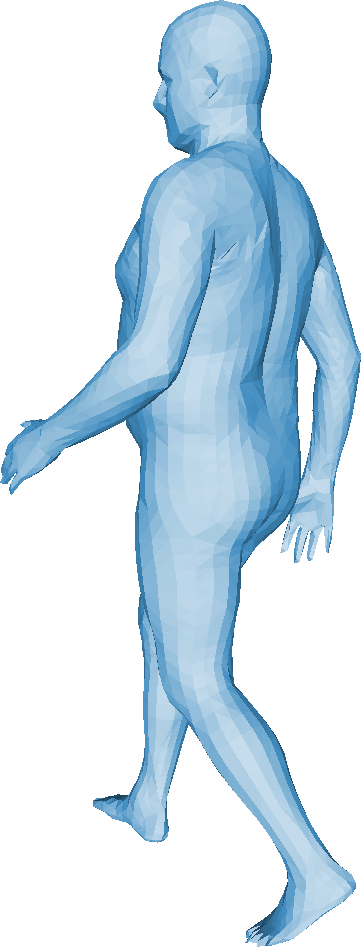} &
    \includegraphics[align=c,scale=0.06]{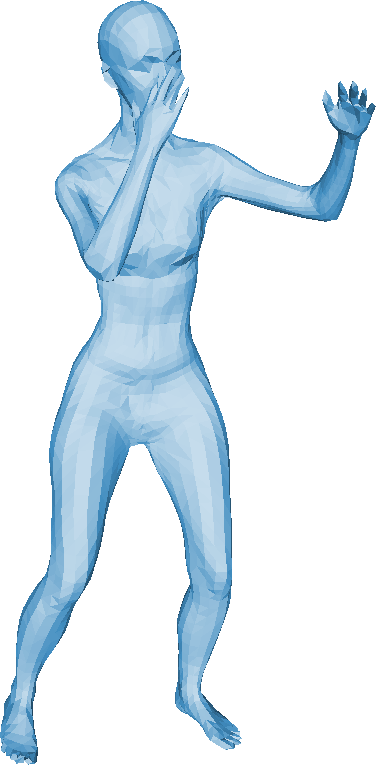} &
    \includegraphics[align=c,scale=0.06]{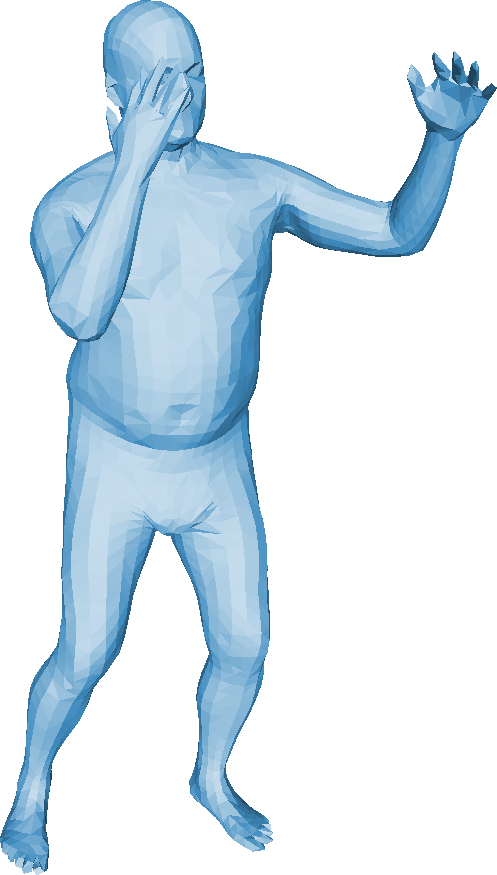} &
    \includegraphics[align=c,scale=0.06]{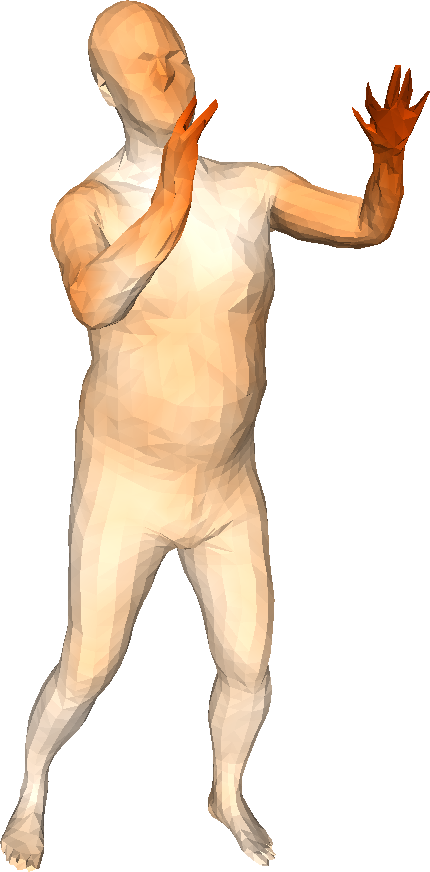} &
    \includegraphics[align=c,scale=0.06]{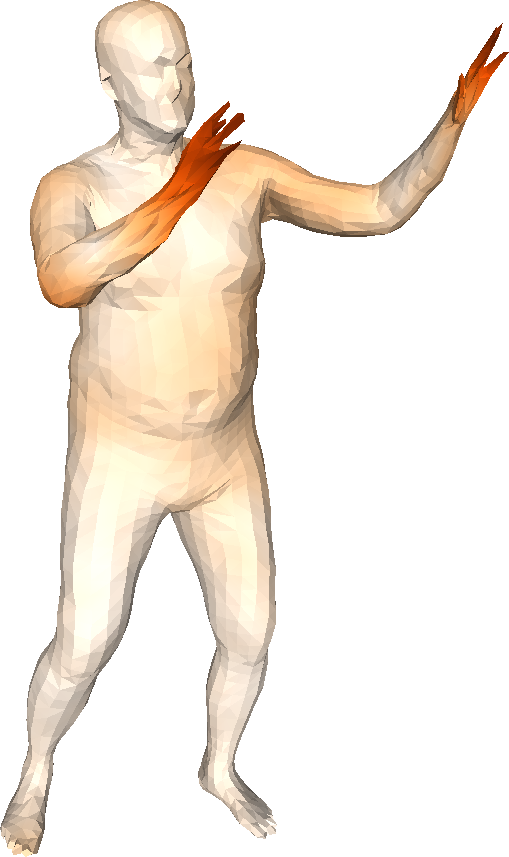} &
    \includegraphics[align=c,scale=0.10]{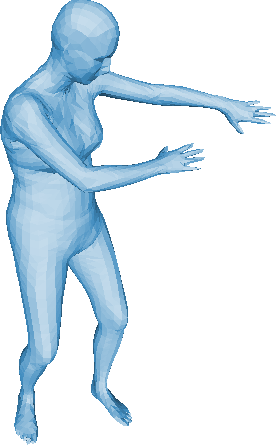} &&
    \includegraphics[align=c,scale=0.10]{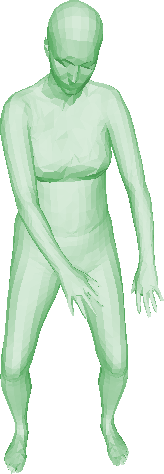}&&
    \includegraphics[align=c,scale=0.10]{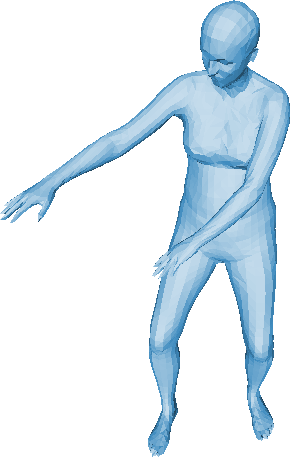} \\ \hline
    \includegraphics[align=c,scale=0.06]{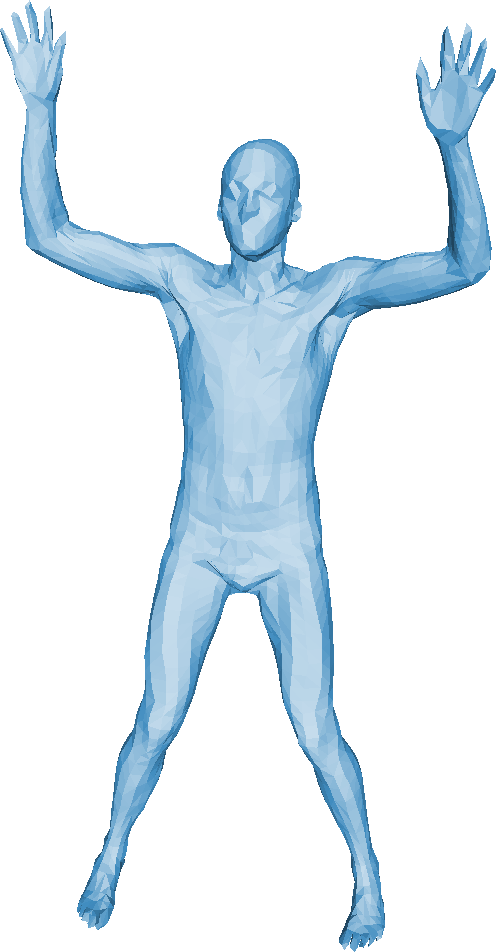} &
    \includegraphics[align=c,scale=0.06]{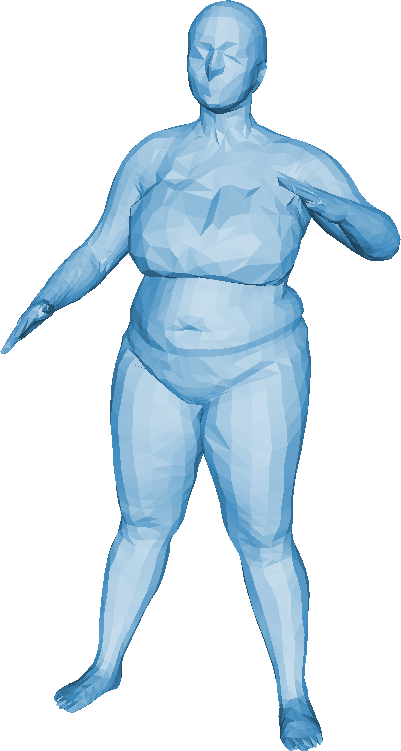} &
    (n/a) &
    \includegraphics[align=c,scale=0.07]{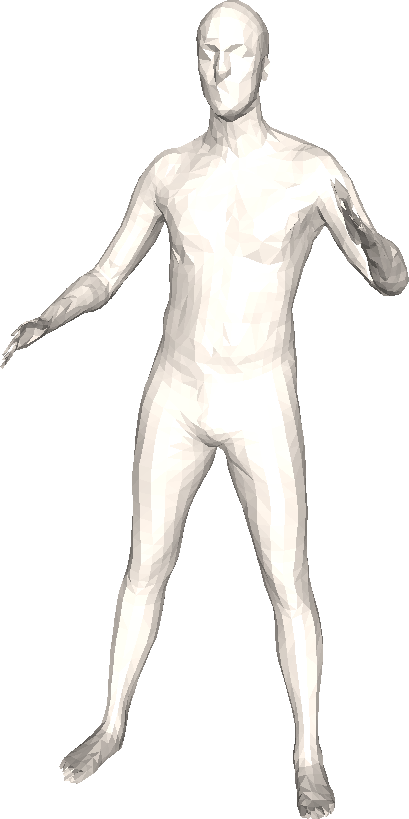} &
    (n/a) &
    \includegraphics[align=c,scale=0.10]{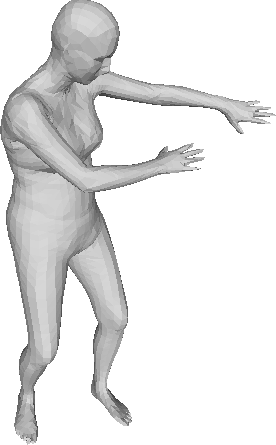} &
    \includegraphics[align=c,scale=0.10]{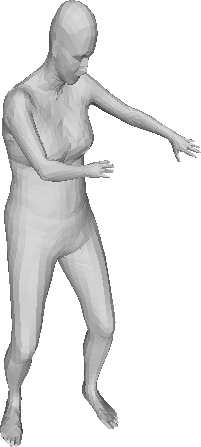} &
    \includegraphics[align=c,scale=0.10]{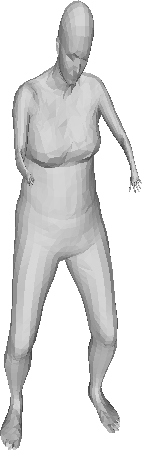} &
    \includegraphics[align=c,scale=0.10]{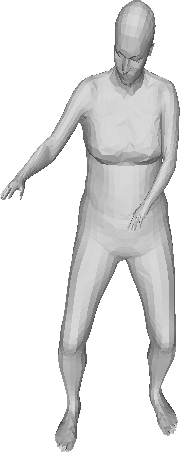} &
    \includegraphics[align=c,scale=0.10]{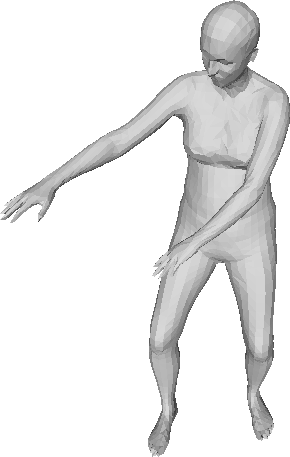} \\
    \end{tabular}
    \caption{\small {\bf Left: Transfer experiment.} The model is given $\mathbf{m}_{\text{diff\_subject}}$ and $\mathbf{m}_{\text{diff\_pose}}$ and combines latent features to predict $\mathbf{m}_{\text{target}}$. The rightmost column shows, for comparison, the result of directly encoding and decoding $\mathbf{m}_{\text{target}}$ itself. The third row shows an example from Faust, which does not have ground truth for pose swapping.
    {\bf Right: Interpolation experiment.} Comparison of latent linear interpolation in feature space (top), compared to extrinsic linear interpolation in $\mathbb{R}^3$ (bottom).
    }
    \label{fig:interpolate plus transfer}
\end{figure}

To explore the local structure of the model's latent space, we took motion sequences and conducted pose interpolation between nearby frames, using linear interpolation in the latent space. The interpolation results in additional semantic detail, such as arm-bending between two poses with almost straight arms. See Fig. \ref{fig:interpolate plus transfer} for a comparison with naive extrinsic linear interpolation of vertex coordinates in $\mathbb{R}^3$, which causes unrealistic mesh deformations.

\section{Conclusion and Future Work}

This paper introduces a disentangled mesh-convolutional VAE. With careful consideration of the supervision and training design, we see that our proposed model can achieve accurate disentanglement while capturing the varied pose and shape properties in large-scale mesh datasets.

Given these promising results, in future work we will explore two directions: 
(1) model improvements by extending our current design by incorporating techniques from alternate generative approaches (e.g. VAE-GAN), and (2) domain transfer to 3D data captured in the wild (e.g. captured with commodity RGBD sensors).

\bibliographystyle{abbrvnat}
\bibliography{bibliography}

\begin{thebibliography}{58}
\providecommand{\natexlab}[1]{#1}
\providecommand{\url}[1]{\texttt{#1}}
\expandafter\ifx\csname urlstyle\endcsname\relax
  \providecommand{\doi}[1]{doi: #1}\else
  \providecommand{\doi}{doi: \begingroup \urlstyle{rm}\Url}\fi

\bibitem[Achille and Soatto(2018)]{achille18jmlr}
A.~Achille and S.~Soatto.
\newblock Emergence of invariance and disentanglement in deep representations.
\newblock \emph{Journal of Machine Learning Research}, 19\penalty0
  (50):\penalty0 1--34, 2018.
\newblock URL \url{http://jmlr.org/papers/v19/17-646.html}.

\bibitem[Achlioptas et~al.(2018)Achlioptas, Diamanti, Mitliagkas, and
  Guibas]{achlioptas18icml}
P.~Achlioptas, O.~Diamanti, I.~Mitliagkas, and L.~Guibas.
\newblock Learning representations and generative models for 3{D} point clouds.
\newblock In \emph{International Conference on Machine Learning, ({ICML})},
  volume~80, pages 40--49, 2018.

\bibitem[Allen et~al.(2006)Allen, Curless, Popovi\'{c}, and
  Hertzmann]{Allen:2006:LCM:1218064.1218084}
B.~Allen, B.~Curless, Z.~Popovi\'{c}, and A.~Hertzmann.
\newblock Learning a correlated model of identity and pose-dependent body shape
  variation for real-time synthesis.
\newblock In \emph{Proceedings of the 2006 ACM SIGGRAPH/Eurographics Symposium
  on Computer Animation}, SCA '06, pages 147--156, Aire-la-Ville, Switzerland,
  Switzerland, 2006. Eurographics Association.
\newblock ISBN 3-905673-34-7.
\newblock URL \url{http://dl.acm.org/citation.cfm?id=1218064.1218084}.

\bibitem[Anguelov et~al.(2005)Anguelov, Srinivasan, Koller, Thrun, Rodgers, and
  Davis]{Anguelov:2005:SSC:1186822.1073207}
D.~Anguelov, P.~Srinivasan, D.~Koller, S.~Thrun, J.~Rodgers, and J.~Davis.
\newblock Scape: Shape completion and animation of people.
\newblock In \emph{ACM SIGGRAPH 2005 Papers}, SIGGRAPH '05, pages 408--416, New
  York, NY, USA, 2005. ACM.
\newblock \doi{10.1145/1186822.1073207}.
\newblock URL \url{http://doi.acm.org/10.1145/1186822.1073207}.

\bibitem[Barron and Malik(2015)]{barron15pami}
J.~T. Barron and J.~Malik.
\newblock Shape, illumination, and reflectance from shading.
\newblock \emph{{IEEE} Trans. Pattern Anal. Mach. Intell.}, 37\penalty0
  (8):\penalty0 1670--1687, 2015.

\bibitem[Barrow and Tenenbaum(1978)]{barrow78cvs}
H.~G. Barrow and J.~M. Tenenbaum.
\newblock Recovering intrinsic scene characteristics from images.
\newblock In \emph{Computer Vision Systems}, 1978.

\bibitem[Bengio et~al.(2013)Bengio, Courville, and Vincent]{bengio13pami}
Y.~Bengio, A.~Courville, and P.~Vincent.
\newblock Representation learning: A review and new perspectives.
\newblock \emph{{IEEE} Trans. Pattern Anal. Mach. Intell.}, 35\penalty0
  (8):\penalty0 1798--1828, 2013.

\bibitem[Bogo et~al.(2017)Bogo, Romero, Pons-Moll, and Black]{dfaust:CVPR:2017}
F.~Bogo, J.~Romero, G.~Pons-Moll, and M.~J. Black.
\newblock Dynamic {FAUST}: {R}egistering human bodies in motion.
\newblock In \emph{IEEE Conf. on Computer Vision and Pattern Recognition
  (CVPR)}, July 2017.

\bibitem[Borosan et~al.(2012)Borosan, Jin, DeCarlo, Gingold, and
  Nealen]{borosan12rig}
P.~Borosan, M.~Jin, D.~DeCarlo, Y.~Gingold, and A.~Nealen.
\newblock Rig{M}esh: Automatic rigging for part-based shape modeling and
  deformation.
\newblock \emph{ACM Transactions on Graphics (TOG)}, 31\penalty0 (6):\penalty0
  198:1--198:9, 2012.

\bibitem[Cashman and Fitzgibbon(2012)]{cashman2012shape}
T.~J. Cashman and A.~W. Fitzgibbon.
\newblock What shape are dolphins? building 3d morphable models from 2d images.
\newblock \emph{IEEE transactions on pattern analysis and machine
  intelligence}, 35\penalty0 (1):\penalty0 232--244, 2012.

\bibitem[Chan et~al.(2018)Chan, Ginosar, Zhou, and
  Efros]{DBLP:journals/corr/abs-1808-07371-dancenow}
C.~Chan, S.~Ginosar, T.~Zhou, and A.~A. Efros.
\newblock Everybody dance now.
\newblock \emph{CoRR}, abs/1808.07371, 2018.
\newblock URL \url{http://arxiv.org/abs/1808.07371}.

\bibitem[Chang et~al.(2015)Chang, Funkhouser, Guibas, Hanrahan, Huang, Li,
  Savarese, Savva, Song, Su, Xiao, Yi, and Yu]{shapenet}
A.~X. Chang, T.~A. Funkhouser, L.~J. Guibas, P.~Hanrahan, Q.~Huang, Z.~Li,
  S.~Savarese, M.~Savva, S.~Song, H.~Su, J.~Xiao, L.~Yi, and F.~Yu.
\newblock Shapenet: An information-rich 3d model repository.
\newblock \emph{CoRR}, abs/1512.03012, 2015.

\bibitem[Chen et~al.(2016)Chen, Duan, Houthooft, Schulman, Sutskever, and
  Abbeel]{infogan}
X.~Chen, Y.~Duan, R.~Houthooft, J.~Schulman, I.~Sutskever, and P.~Abbeel.
\newblock Infogan: Interpretable representation learning by information
  maximizing generative adversarial nets.
\newblock In \emph{Neural Information Processing Systems}, pages 2180--2188,
  2016.

\bibitem[{CMU MoCap}(1999)]{cmu_mocap}
{CMU MoCap}.
\newblock Carnegie-mellon mocap database, 1999.
\newblock URL \url{http://mocap.cs.cmu.edu/}.

\bibitem[de~Bem et~al.(2018)de~Bem, Ghosh, Ajanthan, Miksik, Siddharth, and
  Torr]{de2018dgpose}
R.~de~Bem, A.~Ghosh, T.~Ajanthan, O.~Miksik, N.~Siddharth, and P.~H. Torr.
\newblock Dgpose: Disentangled semi-supervised deep generative models for human
  body analysis.
\newblock \emph{arXiv preprint arXiv:1804.06364}, 2018.

\bibitem[Denton and Birodkar(2017)]{denton17nips}
E.~Denton and V.~Birodkar.
\newblock Unsupervised learning of disentangled representations from video.
\newblock In \emph{Advances in Neural Information Processing Systems}, pages
  4414--4423, 2017.

\bibitem[Dubrovina et~al.(2019)Dubrovina, Xia, Achlioptas, Shalah, and
  Guibas]{dubrovina19arxiv}
A.~Dubrovina, F.~Xia, P.~Achlioptas, M.~Shalah, and L.~J. Guibas.
\newblock Composite shape modeling via latent space factorization.
\newblock \emph{CoRR}, abs/1901.02968, 2019.
\newblock URL \url{http://arxiv.org/abs/1901.02968}.

\bibitem[Freifeld and Black(2012)]{freifeld2012lie}
O.~Freifeld and M.~J. Black.
\newblock Lie bodies: A manifold representation of 3d human shape.
\newblock In \emph{European Conference on Computer Vision}, pages 1--14.
  Springer, 2012.

\bibitem[Gao et~al.(2016)Gao, Lai, Liang, Chen, and Xia]{gao2016efficient}
L.~Gao, Y.-K. Lai, D.~Liang, S.-Y. Chen, and S.~Xia.
\newblock Efficient and flexible deformation representation for data-driven
  surface modeling.
\newblock \emph{ACM Transactions on Graphics (TOG)}, 35\penalty0 (5):\penalty0
  158, 2016.

\bibitem[Gao et~al.(2017)Gao, Lai, Yang, Zhang, Kobbelt, and
  Xia]{gao2017sparse}
L.~Gao, Y.-K. Lai, J.~Yang, L.-X. Zhang, L.~Kobbelt, and S.~Xia.
\newblock Sparse data driven mesh deformation.
\newblock \emph{arXiv preprint arXiv:1709.01250}, 2017.

\bibitem[Gao et~al.(2018)Gao, Yang, Qiao, Lai, Rosin, Xu, and
  Xia]{Gao:2018:AUS:3272127.3275028:DeformationTransfer}
L.~Gao, J.~Yang, Y.-L. Qiao, Y.-K. Lai, P.~L. Rosin, W.~Xu, and S.~Xia.
\newblock Automatic unpaired shape deformation transfer.
\newblock \emph{ACM Trans. Graph.}, 37\penalty0 (6):\penalty0 237:1--237:15,
  Dec. 2018.
\newblock ISSN 0730-0301.
\newblock \doi{10.1145/3272127.3275028}.
\newblock URL \url{http://doi.acm.org/10.1145/3272127.3275028}.

\bibitem[Goodfellow et~al.(2014)Goodfellow, Pouget-Abadie, Mirza, Xu,
  Warde-Farley, Ozair, Courville, and Bengio]{gan}
I.~Goodfellow, J.~Pouget-Abadie, M.~Mirza, B.~Xu, D.~Warde-Farley, S.~Ozair,
  A.~Courville, and Y.~Bengio.
\newblock Generative adversarial nets.
\newblock In \emph{Advances in Neural Information Processing Systems}, pages
  2672--2680, 2014.

\bibitem[Hasler et~al.(2009)Hasler, Stoll, Sunkel, Rosenhahn, and
  Seidel]{hasler2009statistical}
N.~Hasler, C.~Stoll, M.~Sunkel, B.~Rosenhahn, and H.-P. Seidel.
\newblock A statistical model of human pose and body shape.
\newblock In \emph{Computer graphics forum}, volume~28, pages 337--346. Wiley
  Online Library, 2009.

\bibitem[Hinton et~al.(2011)Hinton, Krizhevsky, and Wang]{hinton11icann}
G.~E. Hinton, A.~Krizhevsky, and S.~D. Wang.
\newblock Transforming auto-encoders.
\newblock In \emph{Proceedings of the 21st International Conference on
  Artificial Neural Networks (ICANN)}, pages 44--51, 2011.

\bibitem[Hirshberg et~al.(2012)Hirshberg, Loper, Rachlin, and
  Black]{hirshberg2012coregistration}
D.~A. Hirshberg, M.~Loper, E.~Rachlin, and M.~J. Black.
\newblock Coregistration: Simultaneous alignment and modeling of articulated 3d
  shape.
\newblock In \emph{European Conference on Computer Vision}, pages 242--255.
  Springer, 2012.

\bibitem[Kim and Mnih(2018)]{factorvae}
H.~Kim and A.~Mnih.
\newblock Disentangling by factorising.
\newblock In \emph{International Conference on Machine Learning}, pages
  2649--2658, 2018.

\bibitem[Kingma and Welling(2014)]{vae}
D.~P. Kingma and M.~Welling.
\newblock Auto-encoding variational bayes.
\newblock In \emph{International Conference on Learning Representations
  ({ICLR})}, 2014.

\bibitem[Kingma et~al.(2014)Kingma, Mohamed, Jimenez~Rezende, and
  Welling]{kingma_semisup_vae}
D.~P. Kingma, S.~Mohamed, D.~Jimenez~Rezende, and M.~Welling.
\newblock Semi-supervised learning with deep generative models.
\newblock In \emph{Advances in Neural Information Processing Systems}, pages
  3581--3589, 2014.

\bibitem[Kulkarni et~al.(2015)Kulkarni, Whitney, Kohli, and
  Tenenbaum]{bib:inverse_graphics_network}
T.~D. Kulkarni, W.~F. Whitney, P.~Kohli, and J.~Tenenbaum.
\newblock Deep convolutional inverse graphics network.
\newblock In \emph{Advances in neural information processing systems}, pages
  2539--2547, 2015.

\bibitem[Litany et~al.(2018)Litany, Bronstein, Bronstein, and
  Makadia]{bib:meshvae}
O.~Litany, A.~Bronstein, M.~Bronstein, and A.~Makadia.
\newblock Deformable shape completion with graph convolutional autoencoders.
\newblock In \emph{Proceedings of the IEEE Conference on Computer Vision and
  Pattern Recognition}, pages 1886--1895, 2018.

\bibitem[Liu et~al.(2018)Liu, Wei, Shao, Sheng, Yan, and
  Wang]{liu2018exploring}
Y.~Liu, F.~Wei, J.~Shao, L.~Sheng, J.~Yan, and X.~Wang.
\newblock Exploring disentangled feature representation beyond face
  identification.
\newblock In \emph{CVPR}, 2018.

\bibitem[Locatello et~al.(2018)Locatello, Bauer, Lucic, Gelly, Sch{\"{o}}lkopf,
  and Bachem]{locatello18disentangled}
F.~Locatello, S.~Bauer, M.~Lucic, S.~Gelly, B.~Sch{\"{o}}lkopf, and O.~Bachem.
\newblock Challenging common assumptions in the unsupervised learning of
  disentangled representations.
\newblock \emph{CoRR}, abs/1811.12359, 2018.

\bibitem[Loper et~al.(2015)Loper, Mahmood, Romero, Pons-Moll, and
  Black]{SMPL:2015}
M.~Loper, N.~Mahmood, J.~Romero, G.~Pons-Moll, and M.~J. Black.
\newblock {SMPL}: A skinned multi-person linear model.
\newblock \emph{ACM Trans. Graphics (Proc. SIGGRAPH Asia)}, 34\penalty0
  (6):\penalty0 248:1--248:16, Oct. 2015.

\bibitem[Loper et~al.(2014)Loper, Mahmood, and Black]{Loper:SIGASIA:2014}
M.~M. Loper, N.~Mahmood, and M.~J. Black.
\newblock {MoSh}: Motion and shape capture from sparse markers.
\newblock \emph{ACM Transactions on Graphics, (Proc. SIGGRAPH Asia)},
  33\penalty0 (6):\penalty0 220:1--220:13, Nov. 2014.
\newblock URL \url{http://doi.acm.org/10.1145/2661229.2661273}.

\bibitem[Mathieu et~al.(2016)Mathieu, Zhao, Sprechmann, Ramesh, and
  LeCun]{mathieu16nips}
M.~Mathieu, J.~J. Zhao, P.~Sprechmann, A.~Ramesh, and Y.~LeCun.
\newblock Disentangling factors of variation in deep representation using
  adversarial training.
\newblock In \emph{Advances in Neural Information Processing Systems}, pages
  5041--5049, 2016.

\bibitem[Narayanaswamy et~al.(2017)Narayanaswamy, Paige, van~de Meent,
  Desmaison, Goodman, Kohli, Wood, and Torr]{disentangled_vae}
S.~Narayanaswamy, T.~B. Paige, J.-W. van~de Meent, A.~Desmaison, N.~Goodman,
  P.~Kohli, F.~Wood, and P.~Torr.
\newblock Learning disentangled representations with semi-supervised deep
  generative models.
\newblock In \emph{Advances in Neural Information Processing Systems}, pages
  5925--5935, 2017.

\bibitem[Ntouskos et~al.(2015)Ntouskos, Sanzari, Cafaro, Nardi, Natola, Pirri,
  and Ruiz]{ntouskos2015component}
V.~Ntouskos, M.~Sanzari, B.~Cafaro, F.~Nardi, F.~Natola, F.~Pirri, and M.~Ruiz.
\newblock Component-wise modeling of articulated objects.
\newblock In \emph{Proceedings of the IEEE International Conference on Computer
  Vision}, pages 2327--2335, 2015.

\bibitem[Pauly et~al.(2005)Pauly, Mitra, Giesen, Gross, and
  Guibas]{bib:distortion}
M.~Pauly, N.~J. Mitra, J.~Giesen, M.~Gross, and L.~J. Guibas.
\newblock Example-based 3d scan completion.
\newblock In \emph{Proceedings of the Third Eurographics Symposium on Geometry
  Processing}, SGP '05, Aire-la-Ville, Switzerland, Switzerland, 2005.
  Eurographics Association.
\newblock ISBN 3-905673-24-X.
\newblock URL \url{http://dl.acm.org/citation.cfm?id=1281920.1281925}.

\bibitem[Pishchulin et~al.(2017)Pishchulin, Wuhrer, Helten, Theobalt, and
  Schiele]{pishchulin17pr}
L.~Pishchulin, S.~Wuhrer, T.~Helten, C.~Theobalt, and B.~Schiele.
\newblock Building statistical shape spaces for 3d human modeling.
\newblock \emph{Pattern Recognition}, 2017.

\bibitem[Pons-Moll et~al.(2015)Pons-Moll, Romero, Mahmood, and
  Black]{Dyna:SIGGRAPH:2015}
G.~Pons-Moll, J.~Romero, N.~Mahmood, and M.~J. Black.
\newblock Dyna: A model of dynamic human shape in motion.
\newblock \emph{ACM Transactions on Graphics, (Proc. SIGGRAPH)}, 34\penalty0
  (4):\penalty0 120:1--120:14, Aug. 2015.

\bibitem[Robinette et~al.(1999)Robinette, Daanen, and Paquet]{caesar1999}
K.~Robinette, H.~Daanen, and E.~Paquet.
\newblock The caesar project: a 3-d surface anthropometry survey.
\newblock In \emph{3D Imaging and Modelling}, pages 380 -- 386, 1999.

\bibitem[Sakoe and Chiba(1978)]{dtw}
H.~Sakoe and S.~Chiba.
\newblock Dynamic programming algorithm optimization for spoken word
  recognition.
\newblock \emph{IEEE TRANSACTIONS ON ACOUSTICS, SPEECH, AND SIGNAL PROCESSING},
  26:\penalty0 43--49, 1978.

\bibitem[Schmidhuber(1992)]{schmidhuber92neural}
J.~Schmidhuber.
\newblock Learning factorial codes by predictability minimization.
\newblock \emph{Neural Computation}, 4:\penalty0 863--879, 1992.

\bibitem[Shu et~al.(2017)Shu, Yumer, Hadap, Sunkavalli, Shechtman, and
  Samaras]{NeuralFace2017}
Z.~Shu, E.~Yumer, S.~Hadap, K.~Sunkavalli, E.~Shechtman, and D.~Samaras.
\newblock Neural face editing with intrinsic image disentangling.
\newblock In \emph{IEEE Conference on Computer Vision and Pattern Recognition,
  2017}, 2017.

\bibitem[Shu et~al.(2018)Shu, Sahasrabudhe, G{\"u}ler, Samaras, Paragios, and
  Kokkinos]{shu18eccv}
Z.~Shu, M.~Sahasrabudhe, R.~A. G{\"u}ler, D.~Samaras, N.~Paragios, and
  I.~Kokkinos.
\newblock {Deforming Autoencoders: Unsupervised Disentangling of Shape and
  Appearance}.
\newblock In \emph{The European Conference on Computer Vision, ({ECCV})}, 2018.

\bibitem[Sumner and Popovi\'{c}(2004)]{sumner04sig}
R.~W. Sumner and J.~Popovi\'{c}.
\newblock Deformation transfer for triangle meshes.
\newblock \emph{ACM Trans. Graph.}, 23\penalty0 (3):\penalty0 399--405, 2004.

\bibitem[Tan et~al.(2018)Tan, Gao, Lai, and Xia]{tan18cvpr}
Q.~Tan, L.~Gao, Y.~Lai, and S.~Xia.
\newblock Variational autoencoders for deforming 3d mesh models.
\newblock In \emph{Proceedings of the IEEE Conference on Computer Vision and
  Pattern Recognition}, pages 5841--5850, 2018.

\bibitem[Tran et~al.(2017)Tran, Yin, and Liu]{tran17cvpr}
L.~Tran, X.~Yin, and X.~Liu.
\newblock Disentangled representation learning gan for pose-invariant face
  recognition.
\newblock In \emph{In Proceeding of IEEE Computer Vision and Pattern
  Recognition}, Honolulu, HI, July 2017.

\bibitem[Usman et~al.(2019)Usman, Dufour, Saenko, and
  Bregler]{DBLP:journals/corr/abs-1901-10024:PuppetGAN}
B.~Usman, N.~Dufour, K.~Saenko, and C.~Bregler.
\newblock Puppetgan: Transferring disentangled properties from synthetic to
  real images.
\newblock \emph{CoRR}, abs/1901.10024, 2019.
\newblock URL \url{http://arxiv.org/abs/1901.10024}.

\bibitem[Varol et~al.(2017)Varol, Romero, Martin, Mahmood, Black, Laptev, and
  Schmid]{varol17_surreal}
G.~Varol, J.~Romero, X.~Martin, N.~Mahmood, M.~J. Black, I.~Laptev, and
  C.~Schmid.
\newblock Learning from synthetic humans.
\newblock In \emph{CVPR}, 2017.

\bibitem[Verma et~al.(2018)Verma, Boyer, and Verbeek]{bib:feastnet}
N.~Verma, E.~Boyer, and J.~Verbeek.
\newblock Feastnet: Feature-steered graph convolutions for 3d shape analysis.
\newblock In \emph{Proceedings of the IEEE Conference on Computer Vision and
  Pattern Recognition}, pages 2598--2606, 2018.

\bibitem[Villegas et~al.(2017)Villegas, Yang, Hong, Lin, and
  Lee]{villegas17iclr}
R.~Villegas, J.~Yang, S.~Hong, X.~Lin, and H.~Lee.
\newblock Decomposing motion and content for natural video sequence prediction.
\newblock In \emph{International Conference on Learning Representations}, 2017.

\bibitem[Worrall et~al.(2017)Worrall, Garbin, Turmukhambetov, and
  Brostow]{worrall2017iccv}
D.~E. Worrall, S.~J. Garbin, D.~Turmukhambetov, and G.~J. Brostow.
\newblock Interpretable transformations with encoder-decoder networks.
\newblock In \emph{The IEEE International Conference on Computer Vision
  (ICCV)}, Oct 2017.

\bibitem[Wu et~al.(2016)Wu, Zhang, Xue, Freeman, and Tenenbaum]{3dgan}
J.~Wu, C.~Zhang, T.~Xue, W.~T. Freeman, and J.~B. Tenenbaum.
\newblock Learning a probabilistic latent space of object shapes via 3d
  generative-adversarial modeling.
\newblock In \emph{Advances in Neural Information Processing Systems}, pages
  82--90, 2016.

\bibitem[Yang et~al.(2015)Yang, Reed, Yang, and Lee]{yang2015weaklysupervised}
J.~Yang, S.~E. Reed, M.-H. Yang, and H.~Lee.
\newblock Weakly-supervised disentangling with recurrent transformations for 3d
  view synthesis.
\newblock In \emph{Advances in Neural Information Processing Systems}, pages
  1099--1107, 2015.

\bibitem[Yang et~al.(2014)Yang, Yu, Zhou, Du, Davis, and
  Yang]{yang2014semantic}
Y.~Yang, Y.~Yu, Y.~Zhou, S.~Du, J.~Davis, and R.~Yang.
\newblock Semantic parametric reshaping of human body models.
\newblock In \emph{2014 2nd International Conference on 3D Vision}, volume~2,
  pages 41--48. IEEE, 2014.

\bibitem[Zanfir et~al.(2018)Zanfir, Popa, Zanfir, and
  Sminchisescu]{Zanfir_2018_CVPR}
M.~Zanfir, A.-I. Popa, A.~Zanfir, and C.~Sminchisescu.
\newblock Human appearance transfer.
\newblock In \emph{The IEEE Conference on Computer Vision and Pattern
  Recognition (CVPR)}, June 2018.

\bibitem[Zuffi et~al.(2018)Zuffi, Kanazawa, and Black]{Zuffi:CVPR:2018}
S.~Zuffi, A.~Kanazawa, and M.~J. Black.
\newblock Lions and tigers and bears: Capturing non-rigid, {3D}, articulated
  shape from images.
\newblock In \emph{IEEE Conference on Computer Vision and Pattern Recognition
  (CVPR)}. IEEE Computer Society, 2018.

\end{thebibliography}

\appendix

\renewcommand{\theequation}{S\arabic{equation}}
\renewcommand{\thefigure}{S\arabic{figure}}

\section{Appendix}

\subsection{Network architecture}

Our network architecture used the hidden layer widths indicated in Fig. \ref{fig:widths}. All models were trained in Tensorflow with an Adam optimizer with learning rate decaying exponentially from $10^{-4}$ to $10^{-6}$ over 400k steps. The articulated cylinder models typically converged within 100k steps.

\begin{figure}[h]
    \centering
    \begin{tabular}{|r|l|l|} \hline \renewcommand{\arraystretch}{1.2}
    {\bf Cylinders} & Layers / Units \\ \hline
    Encoder & 16 (1x1 conv), 24, 32, 48, 64 (FeaStNet), mean-pooling \\
    Latent space & 4N = 3N (shape) + 1N (pose), N = 1, 2, 3 \\
    Decoder & $64*|V|$ (FC), 64, 48, 32, 24, 16 (FeaStNet), 3 (1x1 conv) \\ \hline \hline
    {\bf Human shapes} & Layers / Units \\ \hline
    Encoder & 16 (1x1 conv), 32, 64, 96, 128 (FeaStNet), mean-pooling \\
    Latent space & 128 = 16 (shape) + 112 (pose) \\
    Decoder & $128*|V|$ (FC), 128, 96, 64, 32, 16 (FeaStNet), 3 (1x1 conv) \\
    Discriminator & 16 (1x1 conv), 32, 64 (FeaStNet), mean-pooling, then 1 (FC) \\ \hline
    \end{tabular}
    \caption{Network architecture.}
    \label{fig:widths}
\end{figure}

\subsection{Experiments}
\subsubsection{Articulated cylinders}

\begin{figure}[h]
    \centering
    \includegraphics[align=c, height=2in]{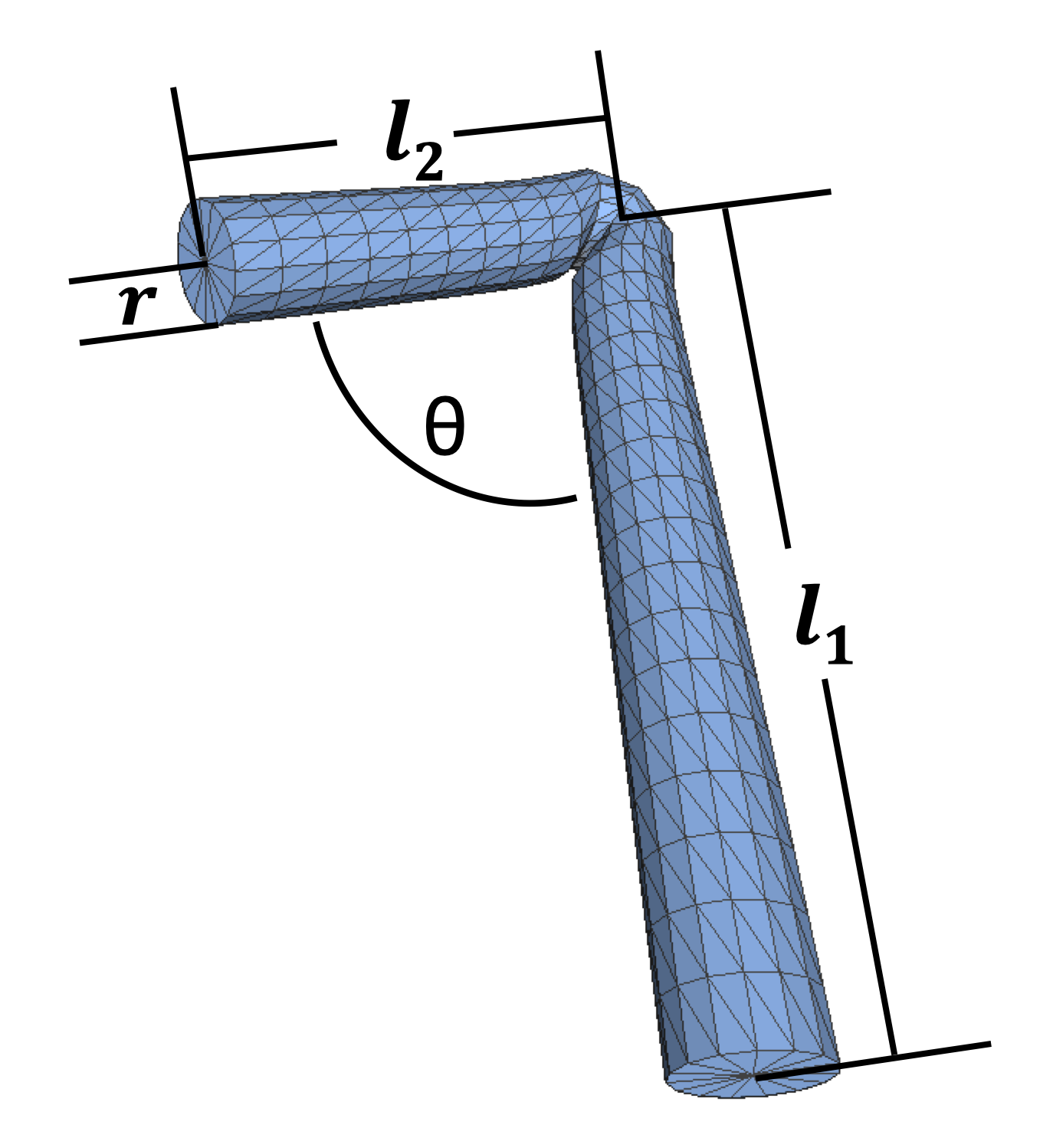}
    \qquad
    \begin{tabular}{|c|c|c|} \hline
        Parameter & Latent swap & Direct \\ \hline
        $\theta$ & 10.25375 & 1.5044 \\ \hline
        $l_1$ & 0.04784 & 0.04711 \\ \hline
        $l_2$ & 0.06322 & 0.0382 \\ \hline
        $r$ & 0.00545 & 0.0051 \\ \hline
    \end{tabular}
    \caption{{\bf Left}: An articulated cylinder. {\bf Right}: Root mean squared error of estimated latent parameters compared to ground truth, for cylinders decoded directly or following latent feature swap.}
    \label{fig:cylinders_rmse}
\end{figure}

We evaluated disentanglement on the articulated cylinders by evaluating the explicit shape and pose parameters from the reconstructed meshes (see Section~4.1 of paper). For test set, we used holdout ranges of parameters, $\theta \in \{[80^\circ,  90^\circ], [140^\circ, 150^\circ]\}, l_1 \in [1.0, 1.25], l2 \in [1.5, 1.75], r \in [0.11, 0.13]$. For a holdout set of $4k$ meshes, we compared latent RMSE (see Fig.~\ref{fig:cylinders_rmse}) and Pearson's correlation (see Fig.~\ref{fig:cylinder2}) for direct reconstructions vs random shape-pose swaps. We noticed the error in shape latent was almost identical, whereas pose latent exhibited a larger error after swap.

\begin{figure}
    \centering
    \begin{tabular}{|c|cccc|} \hline
        Parameter & $\theta_{\text{GT}}$ & $l_1{}_{\text{GT}}$ & $l_2{}_{\text{GT}}$ & $r_{\text{GT}}$ \\ \hline
        $\theta_{\text{est}}$ & 0.99961 & -0.01074 & -0.01690 & -0.00540 \\ \hline
        $l_1{}_{\text{est}}$ & -0.00085 & 0.99973 & - & - \\ \hline
        $l_2{}_{\text{est}}$ & 0.00705 & - & 0.99919 & - \\ \hline
        $r_{\text{est}}$ & -0.02340 & - & - & 0.98127 \\ \hline
    \end{tabular}
    \caption{\small Pearson's correlation between pose ($\theta$) and shape $(l_1, l_2, r)$ parameters for ground truth (columns) and MeshVAE-D direct reconstructions, i.e., without using pose swapping (rows, using estimated parameters). Cross-correlations for pairs of shape features are omitted. Compare to Fig. 1, which shows correlations using estimate latent parameters {\bf after} reconstruction using pose swaps.}
    \label{fig:cylinder2}
\end{figure}

\subsubsection{Discussion on PCA reduction for latent pose generation for human shapes}

\begin{figure}
    \centering
    \includegraphics[width=0.5\linewidth]{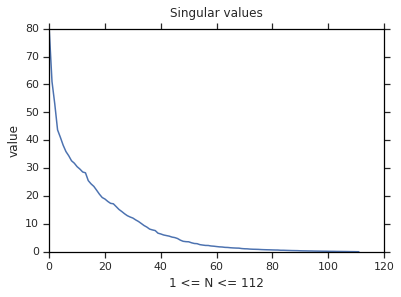}
    \caption{Singular value plot for the latent pose distribution of MeshVAE-D, trained on human shapes. PCA estimated from 220k random training meshes. The top 80 principal components account for $99.994\%$ of the variance.}
    \label{fig: supplementary svd}
\end{figure}

Our model overparametrizes the latent pose vector  (see Fig. \ref{fig: supplementary svd}), offering the possibility of reducing the latent dimensionality after training has finished. For random pose generation, we found it beneficial to first reduce the latent pose space to 80 latent dimensions, by conducting a principal component analysis of the empirical latent distribution (based on the latent pose encodings of approximately 120k meshes from the training set; we found similar results using PCA computed from 40k and 220k meshes) and sampling latent vectors from the top principal components.

A natural question is whether this dimensionality reduction amounts to training with fewer latent features. We compared MeshVAE-D, with 112 latent pose features (called MeshVAE-D-112), to a smaller MeshVAE-D with 80 latent pose features (called MeshVAE-D-80), and found significant impairment to the reconstructed meshes from the smaller model. Specifically, MeshVAE-D-80 had higher mean vertex error for reconstructed meshes ($3.5$cm MVE, compared to $2.7$cm MVE for MeshVAE-D-112), and $65\%$ larger latent embedding variance loss (between meshes with common shape or pose).  
In contrast, applying the latent PCA projection to MeshVAE-D-112 had negligible impacts on mesh reconstruction, increasing the reconstructed mesh MVE by only $0.01$cm (compared to MVE=$2.7$cm for MeshVAE-D-112 without PCA projection). We also note that the PCA on MeshVAE-D-112 transformed the latent pose vectors by $<1\%$ of their norm. Thus, training a model with smaller latent features results in larger reconstruction and disentanglement errors compared to PCA reduction on the latent features of a larger model.

\subsubsection{High-res images of mesh images}
For the reader's enjoyment, we include larger images of the human meshes from Figures 3-7. Meshes colored in blue are originals; grey and white meshes are model outputs.

\begin{figure}
    \centering
    \begin{tabular}{ccc}
    \includegraphics[width=0.3\linewidth]{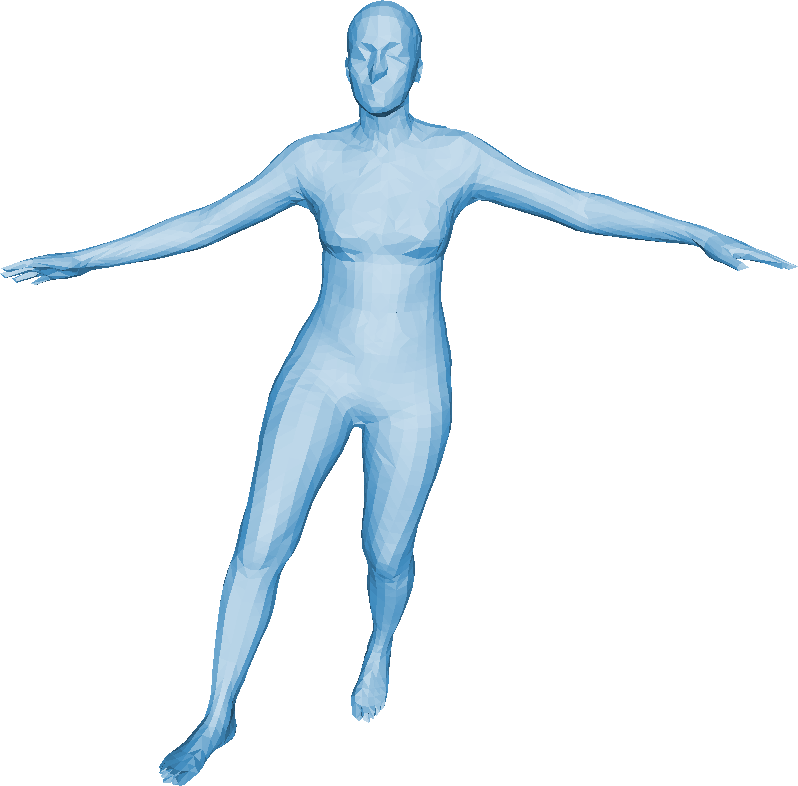} &
    \includegraphics[width=0.3\linewidth]{no_distortion/recon_with_distortion_term.png} &
    \includegraphics[width=0.3\linewidth]{no_distortion/basic_recon.png}
    \end{tabular}
    \caption{Impact of the distortion loss term on reconstructed mesh quality. From left to right: original mesh, MeshVAE-D output, standard MeshVAE output.}
    \label{fig:supplementary distortion and shape recog}
\end{figure}

\begin{figure}
    \centering
    \begin{tabular}{c}
    \includegraphics[height=3.5cm]{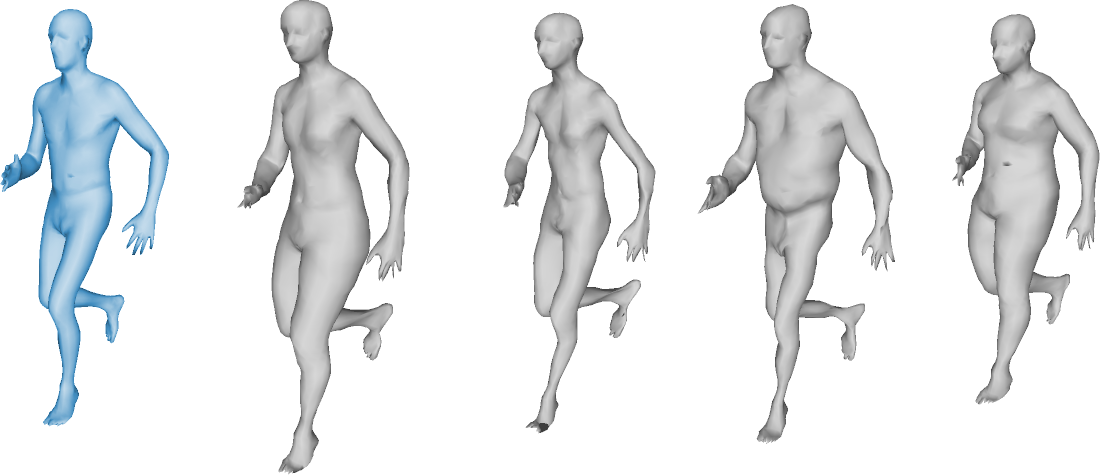} \\
    \includegraphics[height=3.5cm]{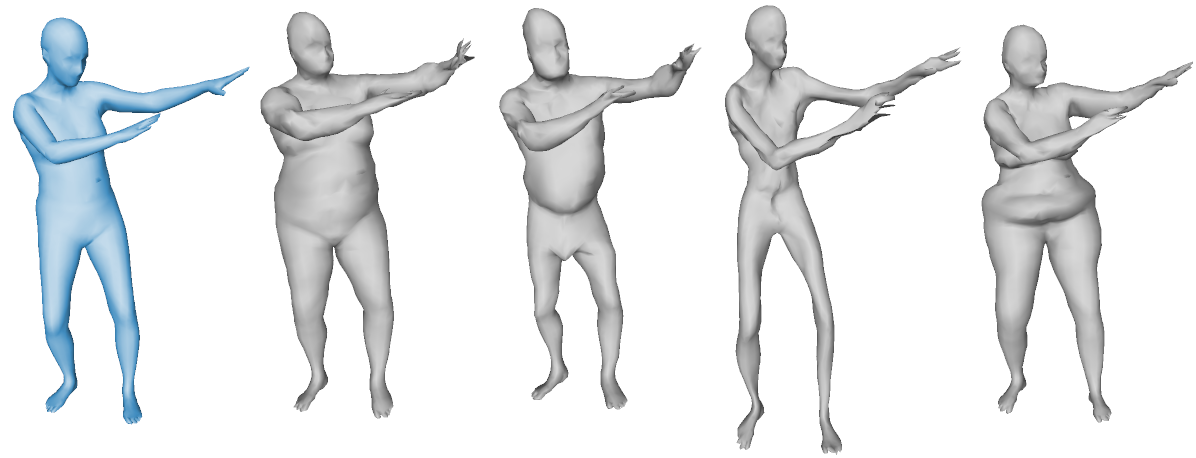} \\
    \includegraphics[height=3.1cm]{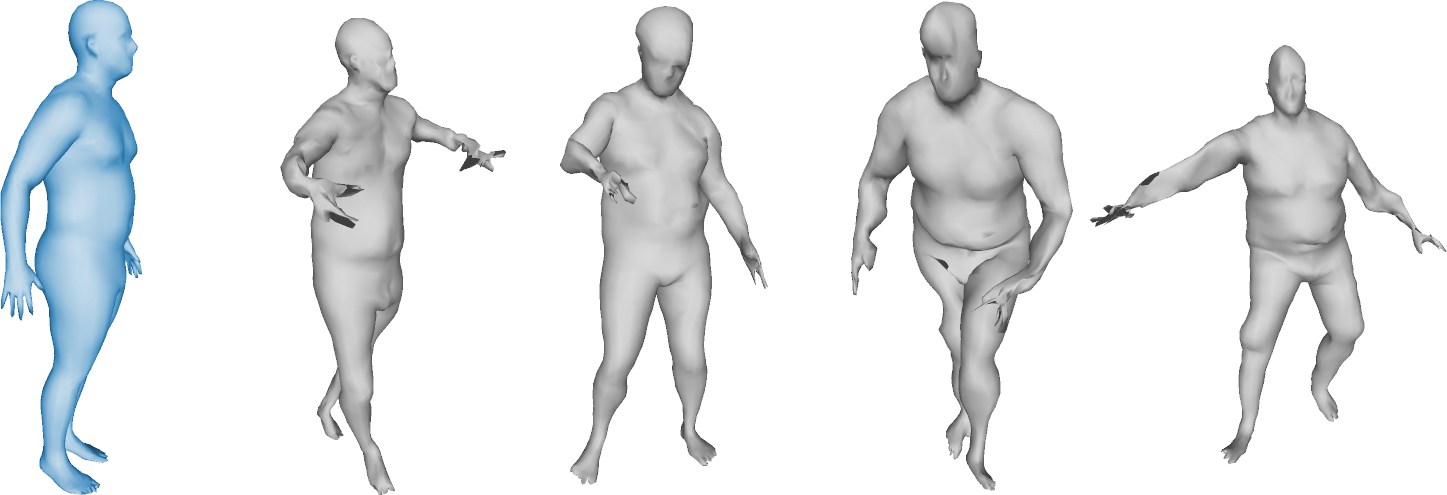} \\
    \includegraphics[height=3.1cm]{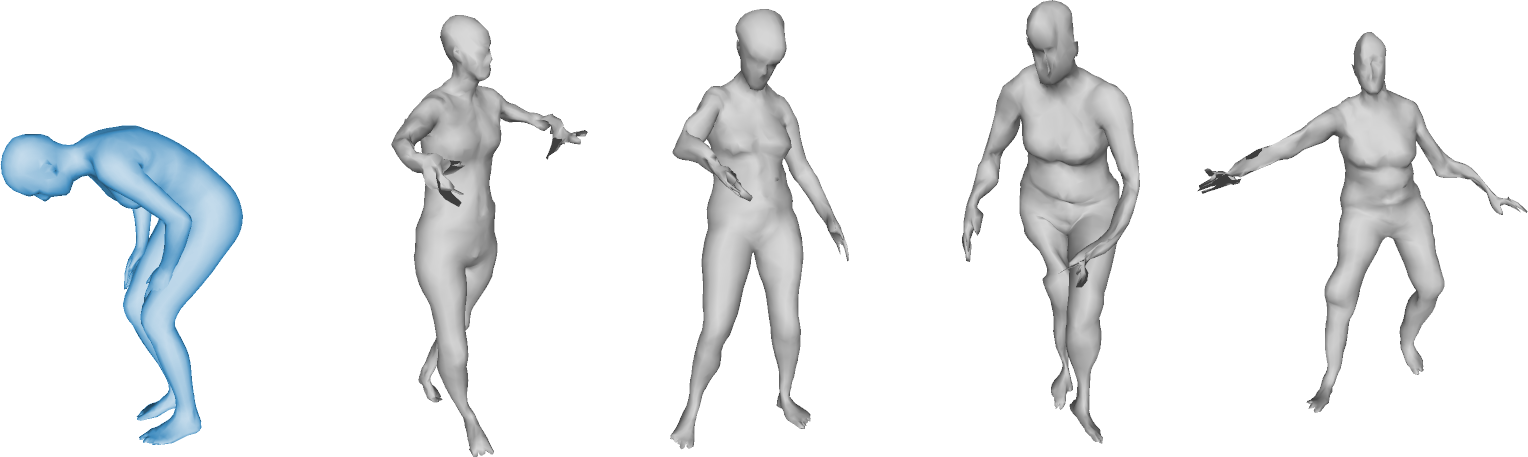}
    \end{tabular}
    \caption{\small Meshes from random shape vectors and fixed real pose encoding (rows 1-2) or vice versa (3-4).}
    \label{fig:supplementary random decodings}
\end{figure}

\begin{figure}
    \centering
    \begin{tabular}{ccc|c|c}
    diff\_pose & diff\_subject & target & combine & direct \\ \hline
    \includegraphics[align=c,scale=0.15]{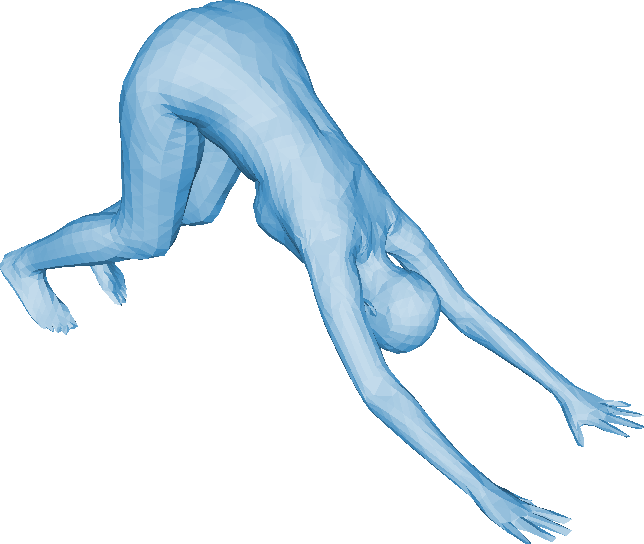} &
    \includegraphics[align=c,scale=0.15]{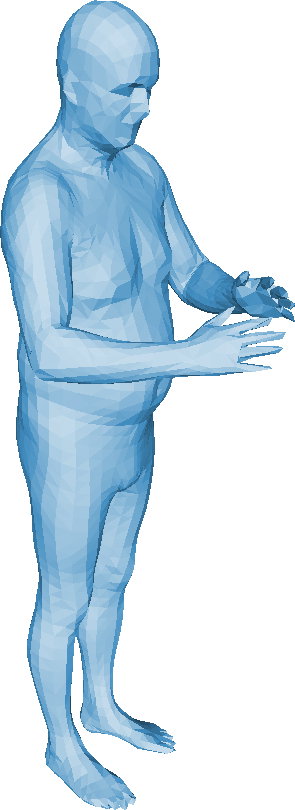} &
    \includegraphics[align=c,scale=0.15]{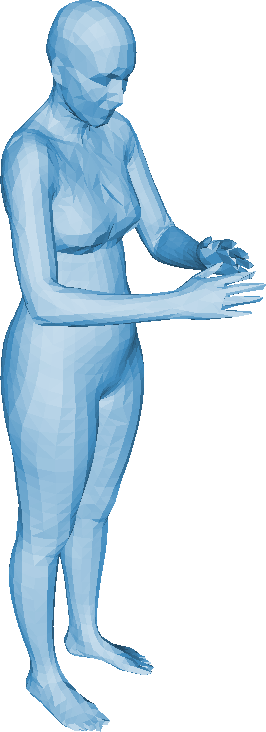} &
    \includegraphics[align=c,scale=0.15]{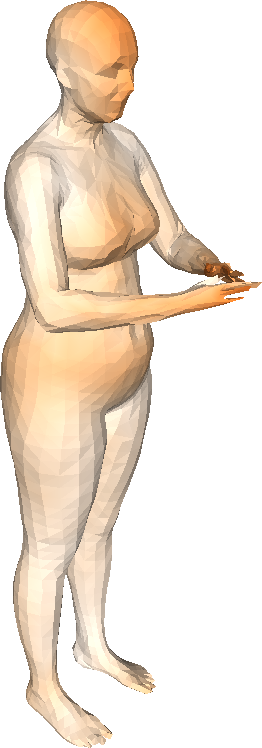} &
    \includegraphics[align=c,scale=0.15]{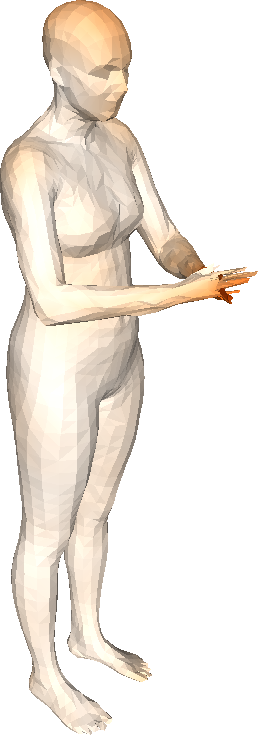} \\ \hline
    \includegraphics[align=c,scale=0.15]{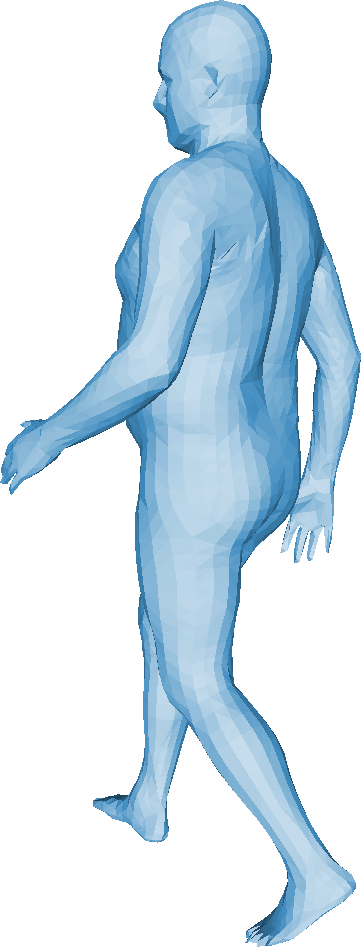} &
    \includegraphics[align=c,scale=0.15]{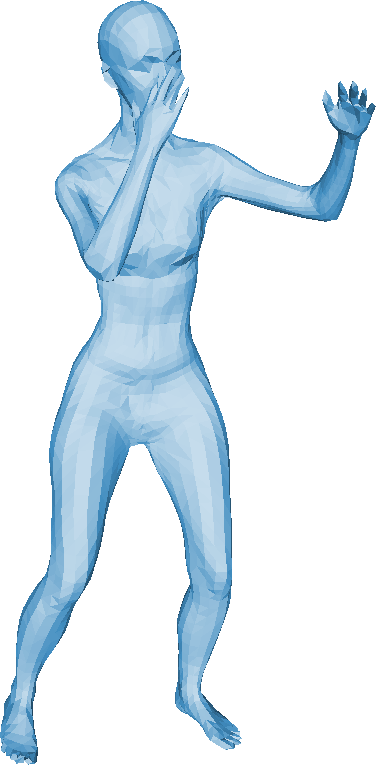} &
    \includegraphics[align=c,scale=0.15]{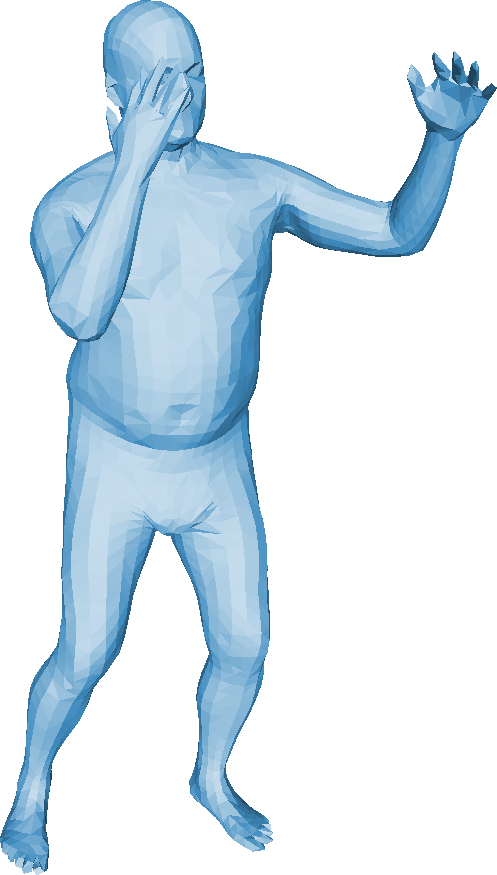} &
    \includegraphics[align=c,scale=0.15]{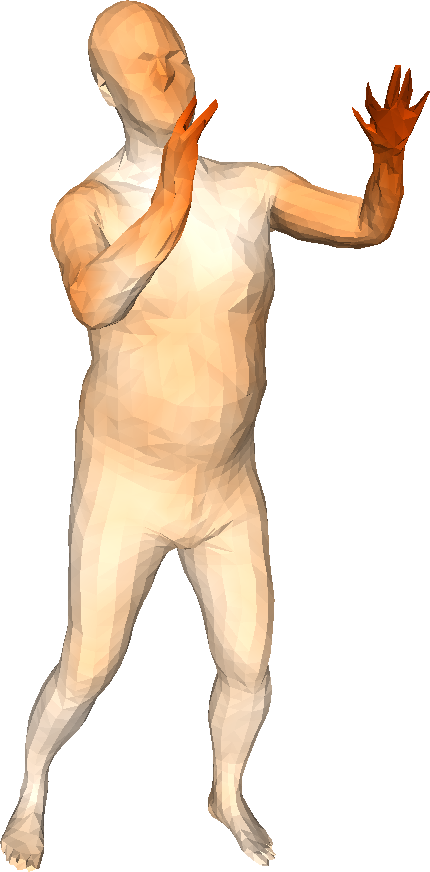} &
    \includegraphics[align=c,scale=0.15]{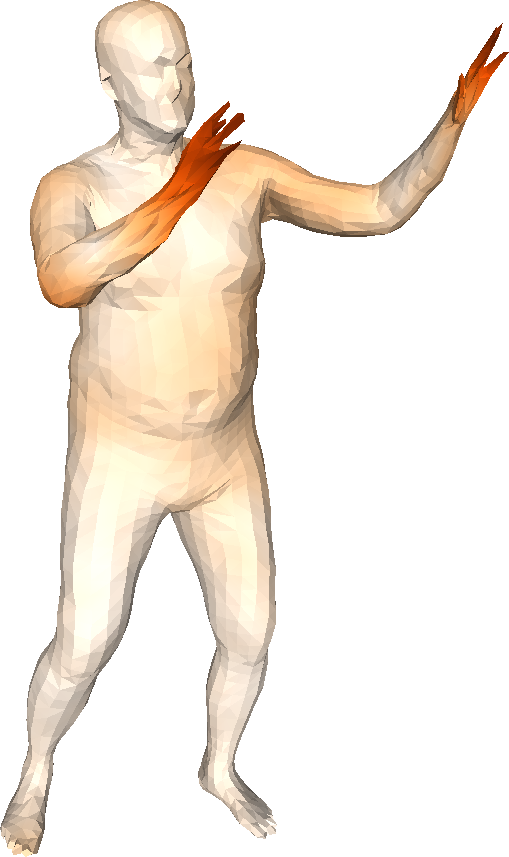} \\ \hline
    \includegraphics[align=c,scale=0.15]{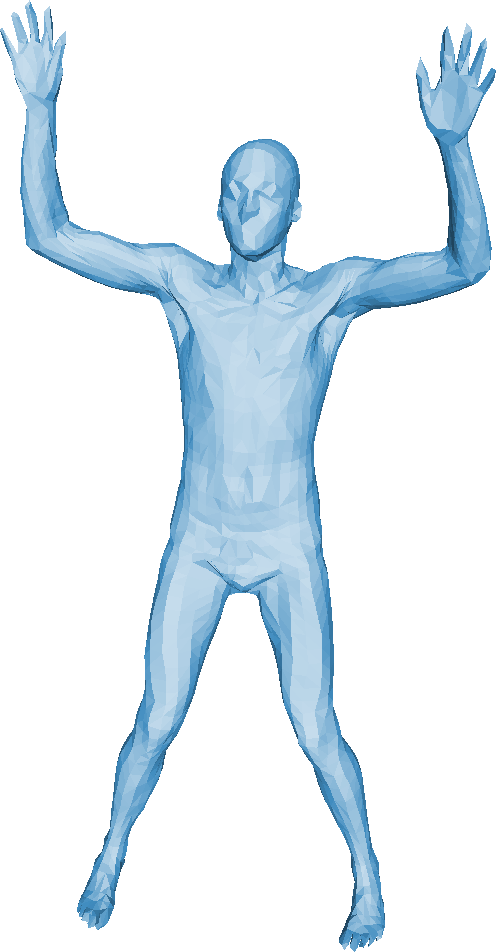} &
    \includegraphics[align=c,scale=0.15]{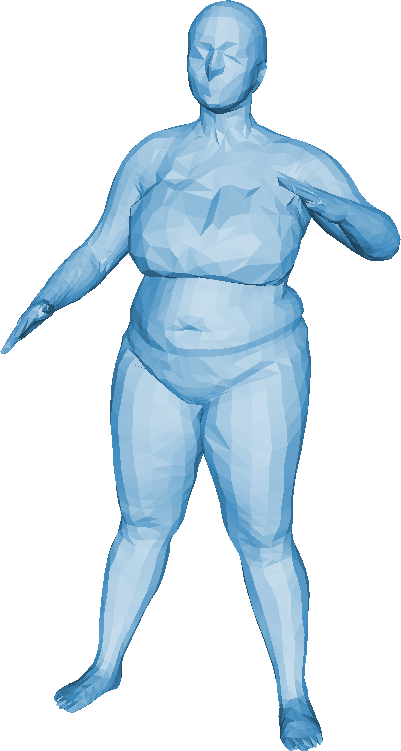} &
    (n/a) &
    \includegraphics[align=c,scale=0.16]{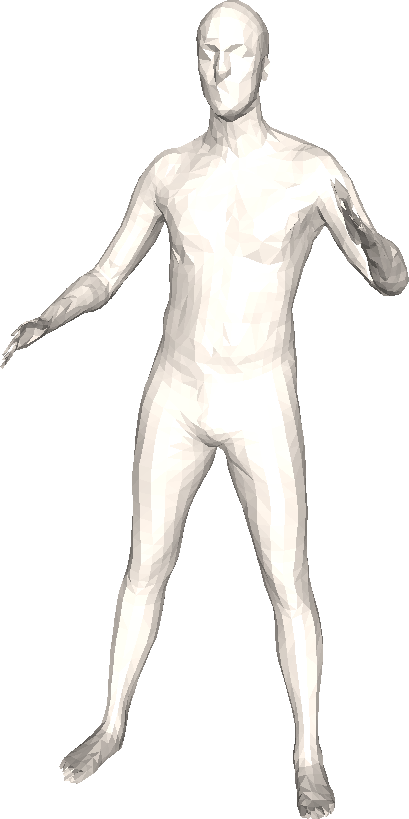} &
    (n/a)
    \end{tabular}
    \caption{ {\bf Transfer experiment.} The model is given $\mathbf{m}_{\text{diff\_subject}}$ and $\mathbf{m}_{\text{diff\_pose}}$ and combines latent features to predict $\mathbf{m}_{\text{target}}$. The rightmost column shows, for comparison, the result of directly encoding and decoding $\mathbf{m}_{\text{target}}$ itself. The third row shows an example from Faust, which does not have ground truth for pose swapping.}
    \label{fig:supplementary pose swap}
\end{figure}

\begin{figure}
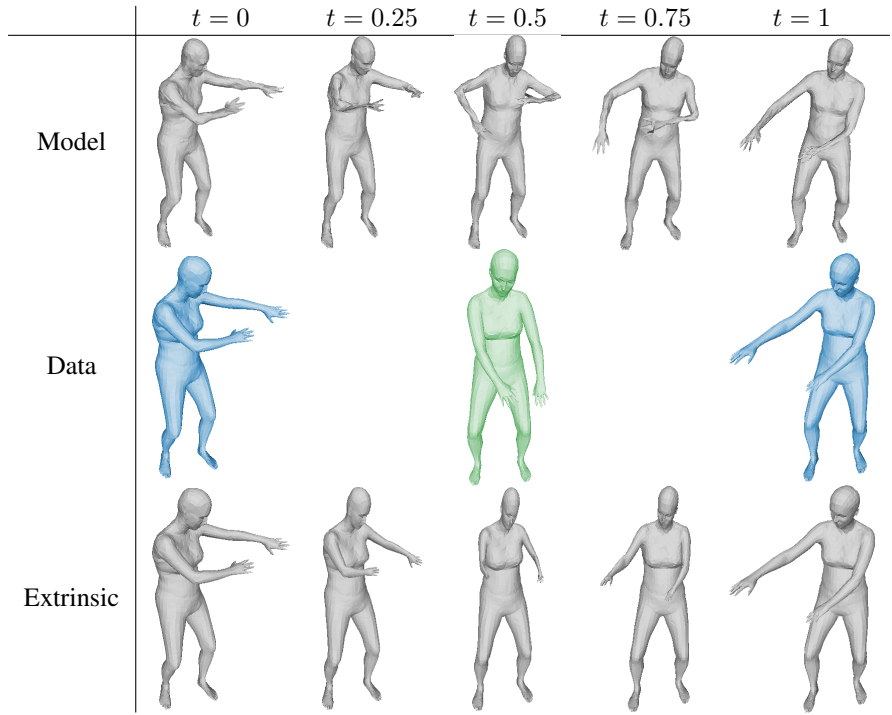

    \centering
    \begin{tabular}{c|ccccc}
    & $t=0$ & $t=0.25$ & $t=0.5$ & $t=0.75$ & $t=1$ \\ \hline
    Model &
    \includegraphics[align=c,scale=0.25]{interpolate/1/latent_00.png} &
    \includegraphics[align=c,scale=0.25]{interpolate/1/latent_10.png} &
    \includegraphics[align=c,scale=0.25]{interpolate/1/latent_20.png} &
    \includegraphics[align=c,scale=0.25]{interpolate/1/latent_30.png} &
    \includegraphics[align=c,scale=0.25]{interpolate/1/latent_40.png} \\
    Data &
    \includegraphics[align=c,scale=0.25]{interpolate/1/orig_0.png} &&
    \includegraphics[align=c,scale=0.25]{interpolate/1/interpolate_mid.png}&&
    \includegraphics[align=c,scale=0.25]{interpolate/1/orig_1.png} \\
    Extrinsic &
    \includegraphics[align=c,scale=0.25]{interpolate/1/r3_00.png} &
    \includegraphics[align=c,scale=0.25]{interpolate/1/r3_10.png} &
    \includegraphics[align=c,scale=0.25]{interpolate/1/r3_20.png} &
    \includegraphics[align=c,scale=0.25]{interpolate/1/r3_30.png} &
    \includegraphics[align=c,scale=0.25]{interpolate/1/r3_40.png} \\
    \end{tabular}
    \caption{{\bf Interpolation experiment.} Interpolating the latent pose encoding, compared to extrinsic linear interpolation in $\mathbb{R}^3$.}
    \label{fig:supplementary interpolate}
\end{figure}

\begin{figure}
    \centering
    \includegraphics[width=\linewidth]{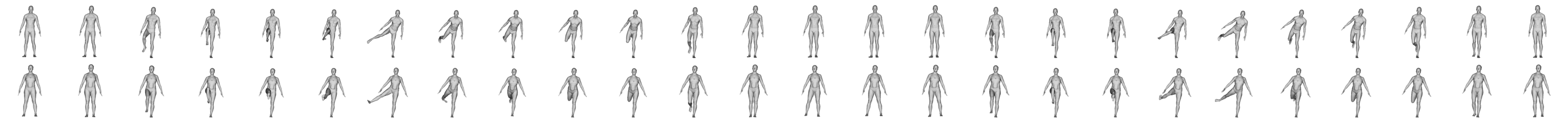} \\
    \includegraphics[width=\linewidth]{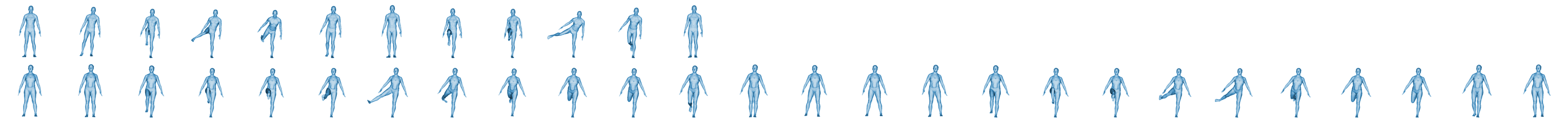}
    \caption{{\bf Top:} Pose synchronization based on the latent pose encoding. Original sequences are shown in blue. Dynamically synchronized sequences shown in gray. Entire sequences shown, in miniature. See next figure for enlarged view.}
    \label{fig:supplementary synchronize}
\end{figure}

\begin{landscape}
\begin{figure}
    \centering{\renewcommand{\arraystretch}{3.0}
    \begin{tabular}{c}
    \raisebox{-.5\height}{\includegraphics[width=\linewidth]{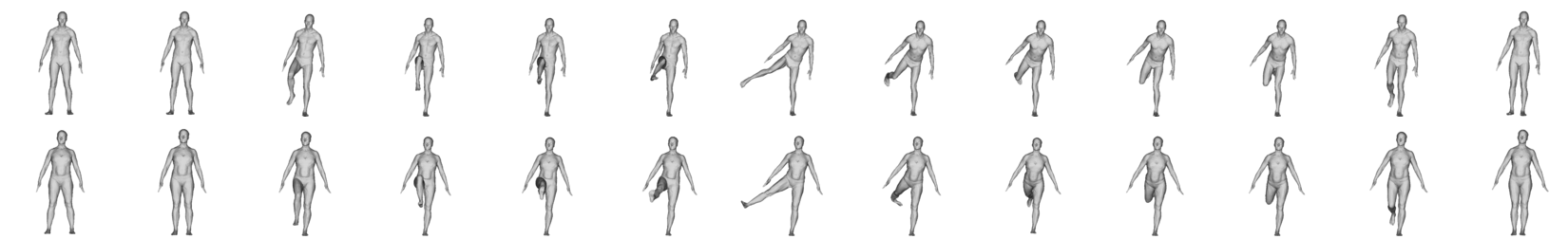}} $\cdots$ \\ \hdashline
    $\cdots$ \raisebox{-.5\height}{\includegraphics[width=\linewidth]{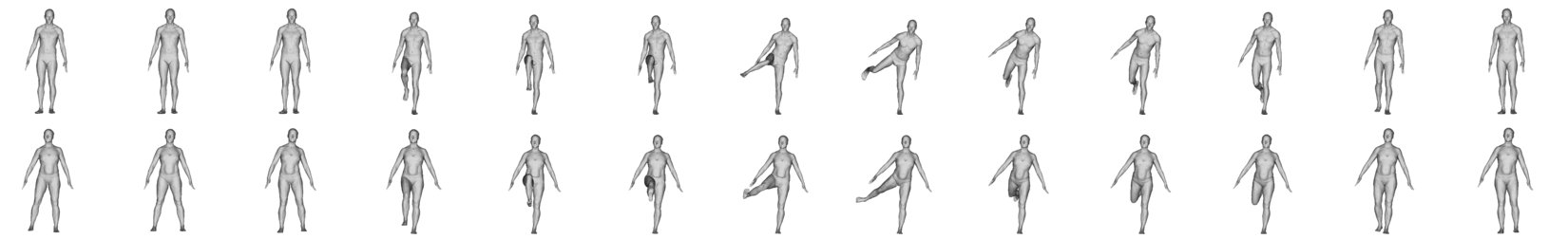}} \\ \hline
    \raisebox{-.5\height}{\includegraphics[width=\linewidth]{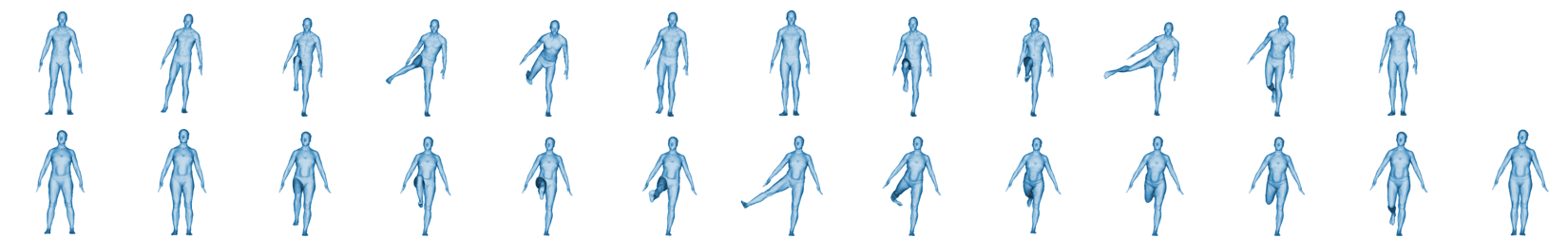}} $\cdots$ \\ \hdashline
    $\cdots$ \raisebox{-.5\height}{\includegraphics[width=\linewidth]{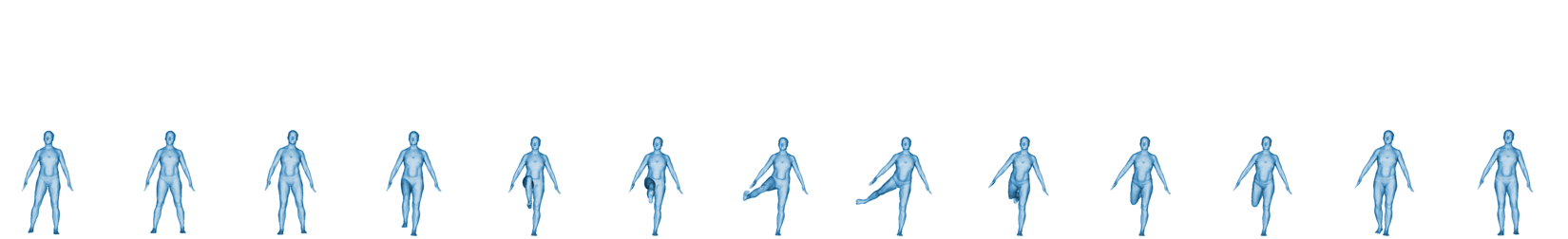}} \\
    \end{tabular}}
    \caption{\small Pose synchronization based on the latent pose encoding. Original sequences are shown in blue. Dynamically synchronized sequences shown in gray.}
    \label{fig:supplementary synchronize full}
\end{figure}
\end{landscape}

\end{document}